\documentclass[11pt,a4paper]{article}
\usepackage[hyperref]{emnlp2020}
\usepackage{times}
\usepackage{latexsym}
\usepackage{amsmath}
\usepackage{amssymb}
\usepackage{hyperref}
\usepackage{graphicx} 
\usepackage{subcaption}
\usepackage{adjustbox}
\usepackage{booktabs}
\usepackage{multirow}
\usepackage{lscape}
\usepackage{xcolor}
\usepackage{colortbl}
\usepackage{xspace}
\usepackage{tabularx}
\usepackage[most]{tcolorbox}

\usepackage{microtype}

\aclfinalcopy % Uncomment this line for the final submission
\def\aclpaperid{2087} %  Enter the acl Paper ID here

\newcommand{\smallsc}[1]{ {\sc \small #1 }}
\newcommand{\datasize}[0]{\textsc{DataSize}}
\newcommand{\curvegrad}[0]{\textsc{CurveGrad}}

\newcommand{\textemb}[0]{\textsc{TextEmb}}
\newcommand{\taskemb}[0]{\textsc{TaskEmb}}
\newcommand{\ttv}[0]{\textsc{Task2Vec}\xspace}

\newcommand{\bvec}[1]{\boldsymbol{#1}}
\definecolor{cellcolor1}{HTML}{D1FED8}
\definecolor{cellcolor2}{HTML}{E6E6FE}
\definecolor{cellcolor3}{HTML}{FCF6C1}
\newcolumntype{t}{>{\columncolor{cellcolor1}}l}
\newcolumntype{a}{>{\columncolor{cellcolor1}}r}

\newcolumntype{u}{l}
\newcolumntype{v}{>{\columncolor{cellcolor3}}l}
\newcolumntype{x}{>{\columncolor{cellcolor1}}c}
\newcolumntype{y}{>{\columncolor{cellcolor2}}c}
\newcolumntype{z}{>{\columncolor{cellcolor3}}c}
\newcommand\blankfootnote[1]{%
  \let\thefootnote\relax\footnotetext{#1}%
  \let\thefootnote\svthefootnote%
}

\title{Exploring and Predicting Transferability across NLP Tasks}

\author{Tu Vu$^1$$^\bigstar$\hspace{9mm}Tong Wang$^2$\hspace{9mm}Tsendsuren Munkhdalai$^2$\hspace{9mm}Alessandro Sordoni$^2$\\\vspace{0.3em}
\textbf{Adam Trischler$^2$\hspace{5mm} Andrew Mattarella-Micke$^3$\hspace{5mm}Subhransu Maji$^1$\hspace{5mm}Mohit Iyyer$^1$} \\ \vspace{0.3em}
  University of Massachusetts Amherst$^1$\hspace{1cm}Microsoft Research Montreal$^2$\hspace{1cm}Intuit AI$^3$\\
  {\tt \{tuvu,smaji,miyyer\}@cs.umass.edu}\\
  {\tt \{tong.wang,tsendsuren.munkhdalai\}@microsoft.com}\\
  {\tt \{alsordo,adam.trischler\}@microsoft.com}\\
  {\tt andrew\_mattarella-micke@intuit.com}}

\date{}

\begin{document}
\maketitle
\begin{abstract}
Recent advances in NLP demonstrate the effectiveness of training large-scale language models and transferring them to downstream tasks. \textit{Can fine-tuning these models on tasks other than language modeling further improve performance?} In this paper, we conduct an extensive study of the transferability between 33 NLP tasks across three broad classes of problems (text classification, question answering, and sequence labeling). Our results show that transfer learning is more beneficial than previously thought, especially when target task data is scarce, and can improve performance even with low-data source tasks that differ substantially from the target task (e.g., part-of-speech tagging transfers well to the DROP QA dataset). We also develop \emph{task embeddings} that can be used to predict the most transferable source tasks for a given target task, and we validate their effectiveness in experiments controlled for source and target data size. Overall, our experiments reveal that factors such as data size, task and domain similarity, and task complexity all play a role in determining transferability.
\blankfootnote{$\bigstar$ Part of this work was done during an internship at Microsoft Research.}
\end{abstract}
\section{Introduction}

With the advent of methods such as ELMo~\citep{MPeters18} and BERT~\citep{JDevlin19}, the dominant paradigm for developing NLP models has shifted to transfer learning: first, pretrain a large language model, and then fine-tune it on the target dataset. Prior work has explored whether fine-tuning on intermediate source tasks before the target task can further improve this pipeline~\citep{JPhang18}, but the conditions for successful transfer remain opaque, and choosing arbitrary source tasks can even adversely impact downstream performance~\cite{AWang19a}. Our work has two main contributions: (1) we perform a large-scale empirical study across 33 different datasets to shed light on the transferability between NLP tasks, and (2) we develop \emph{task embeddings} to predict which source tasks to use for a given target task.

\begin{figure}[t!]
\centering
\includegraphics[width=0.48\textwidth]{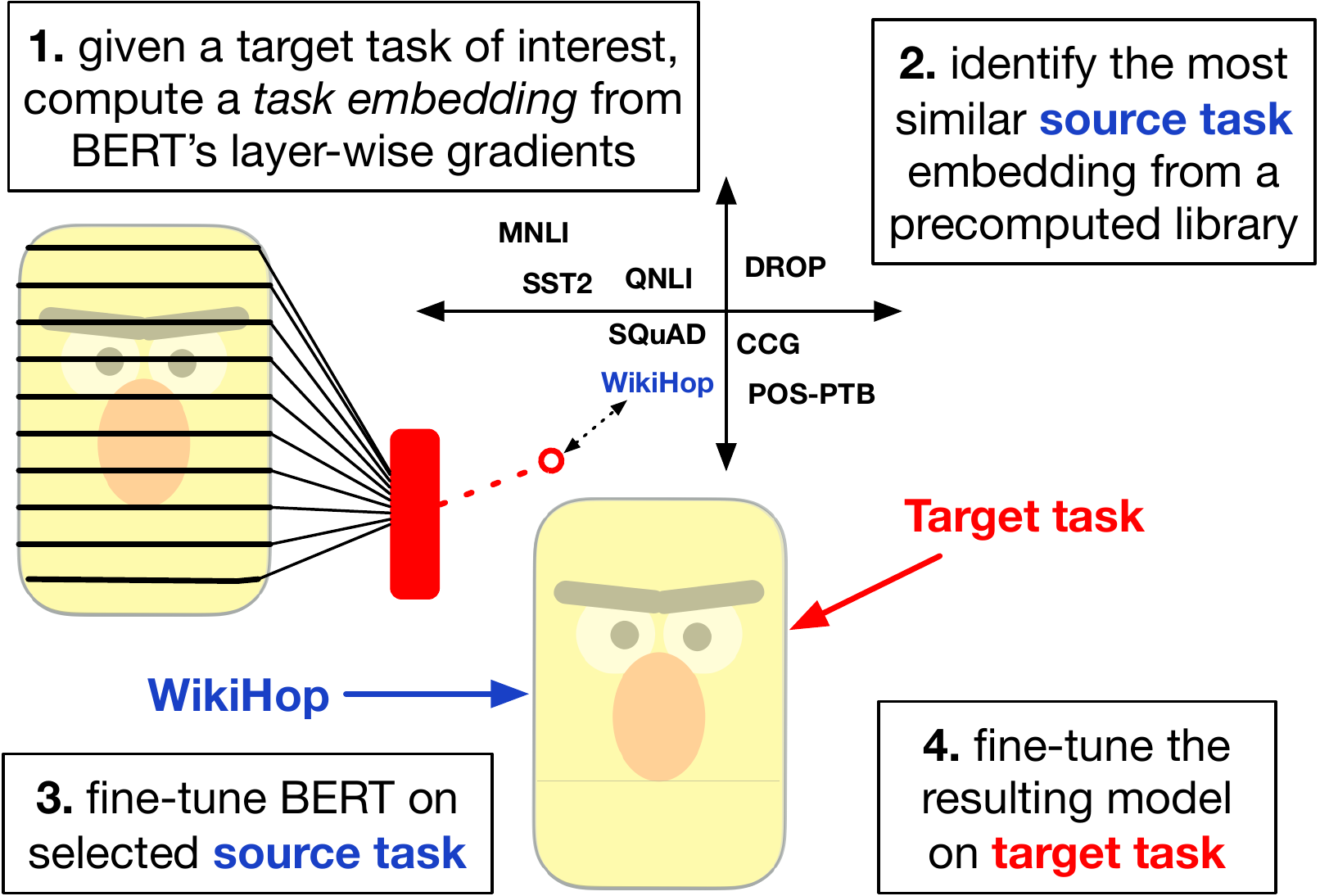}
\caption{A demonstration of our task embedding pipeline. Given a target task, we first compute its task embedding and then identify the most similar source task embedding (in this example, WikiHop) from a precomputed library via cosine similarity. Finally, we perform intermediate fine-tuning of BERT on the selected source task before fine-tuning on the target task.\protect\footnotemark}
\label{fig:taskemb}
\vspace{-2mm}
\end{figure}
\footnotetext{Credit to Jay Alammar for creating the \href{http://jalammar.github.io/illustrated-bert/}{BERT image}.}
Our study includes over 3,000 combinations of tasks and data regimes within and across three broad classes of problems (text classification, question answering, and sequence labeling), which is considerably more comprehensive than prior work~\cite{AWang19c,ATalmor19,NLiu19}. Our results show that transfer learning is more beneficial than previously thought~\cite{AWang19a}, especially for low-data target tasks, and even low-data source tasks that are on the surface very different than the target task can result in transfer gains. While previous work has recommended using the amount of labeled data as a criterion to select source tasks~\citep{JPhang18}, our analysis suggests that the similarity between the source and target tasks and domains are crucial for successful transfer, particularly in data-constrained regimes.

Motivated by these results, we move on to a more practical research question: \textit{given a particular target task, can we predict which source tasks (out of some predefined set) will yield the largest transfer learning improvement, especially in low-data settings?} We address this challenge by learning embeddings of tasks that encode their individual characteristics (Figure~\ref{fig:taskemb}). More specifically, we process all examples from a dataset through BERT and compute a task embedding based on the model's gradients with respect to the task-specific loss, following recent meta-learning work in computer vision~\cite{AAchille19}. We empirically demonstrate the practical value of these task embeddings for selecting source tasks (via simple cosine similarity) that effectively transfer to a given target task. To the best of our knowledge, this is the first work that builds explicit representations of NLP tasks to investigate transferability.

We publicly release our task library, which consists of pretrained models and task embeddings for the 33 NLP tasks we study, along with a codebase that computes task embeddings for new tasks and identifies source tasks that will likely yield positive transferability.\footnote{Library and code available at \href{http://github.com/tuvuumass/task-transferability}{\tt http://github.com/\\tuvuumass/task-transferability}.}
\begin{table}[t!]
\centering
% \footnotesize{
\begin{adjustbox}{max width=0.50\textwidth}
% \scalebox{0.7}{
\begin{tabular}{u r}
\toprule
\multicolumn{1}{l}{\textbf{Task}} & \multicolumn{1}{r}{\textbf{$|$Train$|$}} \\ [0.5ex]
% \midrule
\multicolumn{2}{l}{\emph{text classification/regression (CR)}} \\ [0.5ex] 
SNLI~\cite{SBowman15} & 570K \\
MNLI~\cite{AWilliams18} & 393K \\
QQP~\cite{Quora17} & 364K \\
QNLI~\cite{AWang19a} & 105K \\
SST-2~\cite{RSocher13} &  67K \\
SciTail~\cite{TKhot18} & 27K \\
CoLA~\cite{AWarstadt19} &  8.5K \\
STS-B~\cite{DCer17} & 7K \\
MRPC~\cite{WDolan05} & 3.7K \\
RTE~\cite[][et seq.]{IDagan06} & 2.5K \\
WNLI~\cite{HLevesque11} & 634 \\
\midrule 
\multicolumn{2}{l}{\emph{question answering (QA)}} \\ [0.5ex]
SQuAD-2~\cite{PRajpurkar18} & 162K \\
NewsQA~\cite{ATrischler17} & 120K \\
HotpotQA~\cite{ZYang18} & 113K \\
SQuAD-1~\cite{PRajpurkar16} & 108K \\
DuoRC-p~\cite{ASaha18} & 100K \\
DuoRC-s~\cite{ASaha18} & 86K \\
DROP~\cite{DDua19} & 77K \\
WikiHop~\cite{JWelbl18} & 51K \\
BoolQ~\cite{CClark19} & 16K \\
ComQA~\cite{AAbujabal19} & 11K \\
CQ~\cite{JBao16} & 2K \\
\midrule
\multicolumn{2}{l}{\emph{sequence labeling (SL)}} \\ [0.5ex] 
ST~\cite{JBjerva16} & 43K \\
CCG~\cite{JHockenmaier07} & 40K \\
Parent~\cite{NLiu19} & 40K \\
GParent~\cite{NLiu19} & 40K \\
GGParent~\cite{NLiu19} & 40K \\
POS-PTB~\cite{MMarcus93} & 38K \\
GED~\cite{HYannakoudakis11} & 29K \\
NER~\cite{TKSang03} & 14K \\
POS-EWT~\cite{NSilveira14} & 13K \\
Conj~\cite{JFicler16} & 13K \\
Chunk~\cite{TKSang00} & 9K \\
\bottomrule
\end{tabular}
% }
\end{adjustbox}
\caption{Datasets used in our experiments, grouped by task class and sorted by training dataset size.}
\label{table:datasets}
% }
\vspace*{-2mm}
\end{table}
\section{Exploring task transferability}
\label{sec:transfer}

To shed light on the transferability between different NLP tasks,\footnote{We define a \emph{task} as a (dataset, objective function) pair.} we perform an empirical study with 33 tasks across three broad classes of problems: text classification/regression (CR), question answering (QA), and sequence labeling (SL).\footnote{We divide tasks into classes based on how they are modeled; there is considerable in-class linguistic diversity.}  In each experiment, we follow the STILTs pipeline of~\citet{JPhang18} by taking a pretrained BERT model,\footnote{We use \href{https://github.com/google-research/bert}{BERT-Base Uncased}, which has 12 layers, 768-d hidden size, 12 heads, and 110M total parameters.} fine-tuning it on an intermediate \emph{source} task, and then fine-tuning the resulting model on a \emph{target} task. We explore in-class and out-of-class transfer in both data-rich and data-constrained regimes and demonstrate that positive transfer can occur in a more diverse array of settings than previously thought~\cite{AWang19a}.

\subsection{Experimental setup}
We denote a dataset $D = \{(x^i, y^i)\}_{i=1}^{n}$, with $n$ total examples of inputs $x$ and associated outputs $y$. Each input $x$, which can be either a single text or a concatenation of multiple text segments (e.g., a question-passage pair), is represented as:
\[
\textsc{[cls]} \; w^1_1 \; w^1_2 \;\ldots w^1_{L_1} \; \textsc{[sep]} \; w^2_1 \; w^2_2 \;\ldots w^2_{L_2},
\]
% % use the following line to reduce space if necessary
% \\[1mm]\hspace*{4mm}$
% \textsc{[cls]} \; w^1_1 \; w^1_2 \;\ldots w^1_{L_1} \; \textsc{[sep]} \; w^2_1 \; w^2_2 \;\ldots w^2_{L_2}$,\\[1mm] 
where $w^i_j$ is token $i$ of the $j^{\text{th}}$ segment, \textsc{[cls]} is a special symbol for classification output, and \textsc{[sep]} is a special symbol to separate any text segments if they exist. Finally, each task is solved by applying a classification layer over either the final \textsc{[cls]} token representation (for CR) or the entire sequence of final layer token representations (for QA or SL). 
For both stages of fine-tuning, we follow~\citet{JDevlin19} by backpropagating into all model parameters for a fixed number of epochs.\footnote{We fine-tune all CR and QA tasks for three epochs, and SL tasks for six epochs, using the Transformers library~\cite{TWolf19} and its recommended hyperparameters.} While individual task performance can likely be further improved with more involved hyperparameter tuning for each experimental setting, we standardize hyperparameters across each of the three classes to cut down on computational expense, following prior work~\citep{JPhang18,AWang19a}.

\subsubsection{Datasets \& data regimes}
\label{sec:data_regimes}
Table~\ref{table:datasets} lists the 33 datasets in our study.\footnote{Appendix~\ref{appendix:a1} contains more details about dataset characteristics and their associated evaluation metrics.} We select these datasets by mostly following prior work:  nine of the eleven CR tasks come from the GLUE benchmark~\cite{AWang19a}; all eleven QA tasks are from the MultiQA repository~\cite{MultiQA19}; and all eleven SL tasks were used by~\citet{NLiu19}. We consider all possible pairs of source and target datasets;\footnote{All experiments conducted on a GPU cluster operating on renewable energy.} while some training datasets contain overlapping examples (e.g., SQuAD-1 and 2), we evaluate our models on target development sets, which do not contain overlap.

For each (source, target) dataset pair, we perform transfer experiments in three data regimes to examine the impact of data size on  \smallsc{source~$\rightarrow$~target} transfer: \smallsc{Full $\rightarrow$ Full}, \smallsc{Full $\rightarrow$ Limited}, and \smallsc{Limited $\rightarrow$ Limited.} In the \smallsc{Full} training regime, all training data for the associated task is used for fine-tuning. In the \smallsc{Limited} setting, we artificially limit the amount of training data by randomly selecting 1K training examples without replacement, following~\citet{JPhang18}; since fine-tuning BERT can be unstable on small datasets~\cite{JDevlin19}, we perform 20 random restarts for each experiment and report the mean.\footnote{See Appendix~\ref{appendix:b} for variance statistics. We resample 1K examples for each restart; for tasks with fewer than 1K training examples, we use the full training dataset.}

We measure the impact of transfer learning by computing the \emph{relative transfer gain} given a source task $s$ and target task $t$. More concretely, if a baseline model that is directly fine-tuned on the target dataset (without any intermediate fine-tuning) achieves a performance of $p_{t}$, while a transferred model achieves a performance of $p_{s \rightarrow t}$, the relative transfer gain is defined as: $g_{s\rightarrow t} = \dfrac{p_{s \rightarrow t} - p_{t}}{p_t}.$
\begin{table}[t!]
\centering
\begin{adjustbox}{max width=0.5\textwidth}
\begin{tabular}{ l r @{\hspace*{7mm}} r @{\hspace*{7mm}} r}
\toprule
\multicolumn{4}{l}{\smallsc{Full $\rightarrow$ Full}} \\[0.5ex]
\multicolumn{1}{l}{\footnotesize{$\downarrow$src,tgt$\rightarrow$ }} & \multicolumn{1}{l}{CR} & \multicolumn{1}{l}{QA} & \multicolumn{1}{l}{SL} \\
CR & \cellcolor{cellcolor1} 6.3 \footnotesize(11) & 3.4 \footnotesize(10) & 0.3 \footnotesize(10) \\
QA & 3.2 \footnotesize(10) & \cellcolor{cellcolor1} 9.5 \footnotesize(11) & 0.3 \footnotesize(9) \\
SL & 5.3 \footnotesize(8) & 2.5 \footnotesize(10) & \cellcolor{cellcolor1} 0.5 \footnotesize(11) \\
\midrule
\multicolumn{4}{l}{\smallsc{Full $\rightarrow$ Limited}} \\[0.5ex]
& \multicolumn{1}{l}{CR} & \multicolumn{1}{l}{QA} & \multicolumn{1}{l}{SL} \\
CR & \cellcolor{cellcolor1} 56.9 \footnotesize(11) & 36.8 \footnotesize(10) & 2.0 \footnotesize(10)\\
QA & 44.3 \footnotesize(11) & \cellcolor{cellcolor1} 63.3 \footnotesize(11) & 5.3 \footnotesize(11)\\
SL & 45.6 \footnotesize(11) & 39.2 \footnotesize(6) & \cellcolor{cellcolor1} 20.9 \footnotesize(11)\\
\midrule
\multicolumn{4}{l}{\smallsc{Limited $\rightarrow$ Limited}} \\[0.5ex]
& \multicolumn{1}{l}{CR} & \multicolumn{1}{l}{QA} & \multicolumn{1}{l}{SL} \\
CR & \cellcolor{cellcolor1} 23.7 \footnotesize(11) & 7.3 \footnotesize(11) & 1.1 \footnotesize(11)\\
QA & 37.3 \footnotesize(11) & \cellcolor{cellcolor1} 49.3 \footnotesize(11) & 4.2 \footnotesize(11) \\
SL & 29.3 \footnotesize(10) & 30.0 \footnotesize(8) & \cellcolor{cellcolor1} 10.2 \footnotesize(11)\\
\bottomrule
\end{tabular}
\end{adjustbox}
\caption{A summary of our transfer results for each combination of the three task classes in the three data regimes. Each cell represents the relative gain of the \emph{best} source task in the source class (row) for a given target task, averaged across all of target tasks in the target class (column). In parentheses, we additionally report the number of target tasks (out of 11) for which at least one source task results in a positive transfer gain. The \colorbox{cellcolor1}{diagonal cells} indicate in-class transfer.
}
\label{table:transfer}
\vspace*{-2mm}
\end{table}

\subsection{Analyzing the transfer results}
Table~\ref{table:transfer} contains the results of our transfer experiments across each combination of classes and data regimes.\footnote{See Appendix~\ref{appendix:b} for tables for each individual task.} In each cell, we first compute the transfer gain of the \emph{best} source task for each target task in a particular class, and then average across all target tasks in the same class. We summarize our findings as follows: 
\begin{itemize}
    \itemsep0em 
    \item Contrary to prior belief, transfer gains are possible even when the source dataset is small.
    \item Out-of-class transfer succeeds in many cases, some of which are unintuitive.
    \item Factors other than source dataset size, such as the similarity between source and target tasks, matter more in low-data regimes.
\end{itemize}
In the rest of this section, we analyze each of these three findings in more detail.
\begin{figure*}[ht!]
\centering
\includegraphics[width=\textwidth]{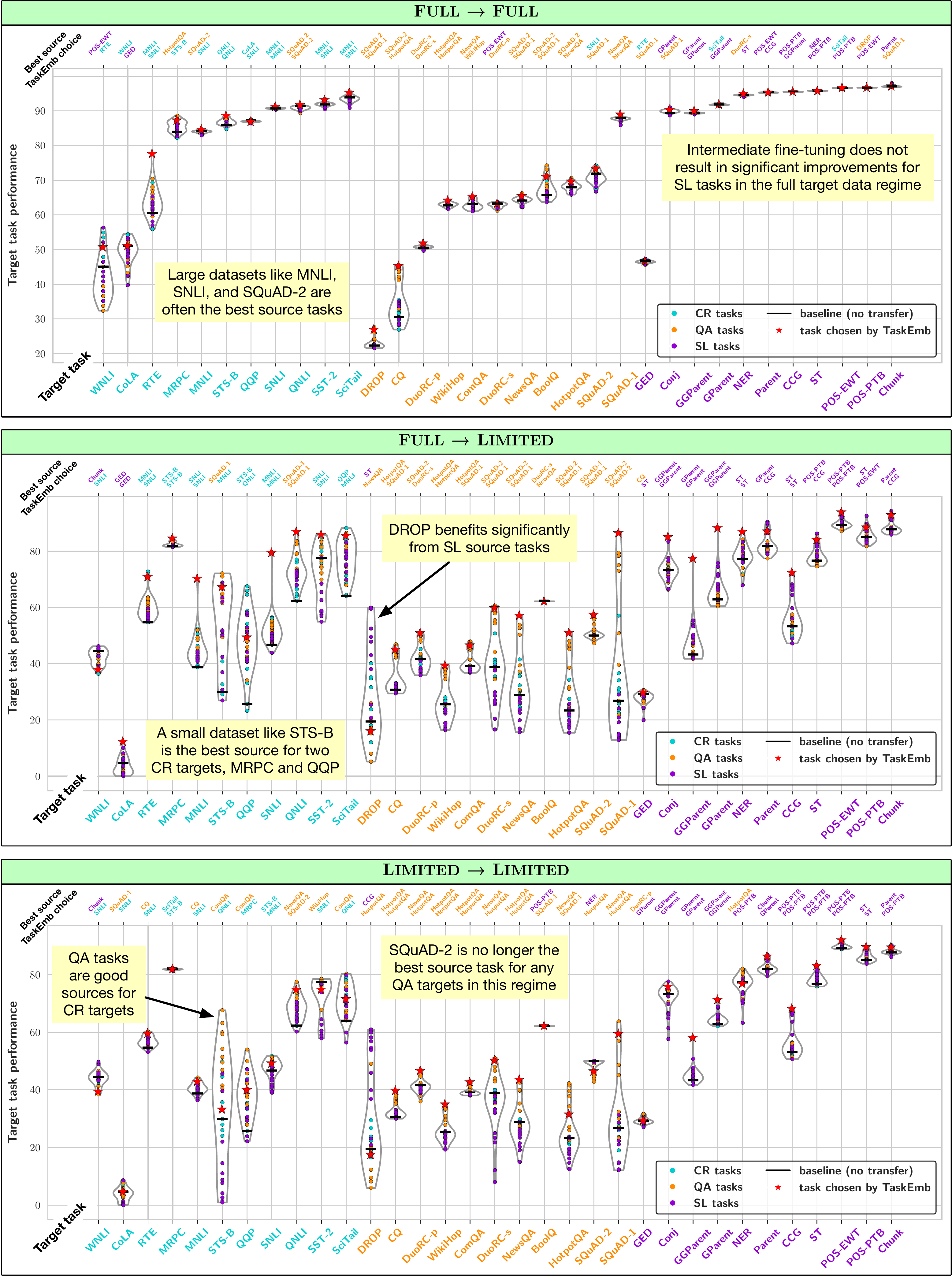}
\vspace*{-2mm}
\caption{In these plots (best viewed in zoom with color), each violin corresponds to a target task in the specified data regime. Each point inside a violin represents an individual source task; its color denotes task class, and its y-coordinate denotes target task performance after transfer. Above each violin, we provide the \emph{best} source task (highest point within the violin) and \taskemb's top-ranked source task (the red star). 
The horizontal black line in each violin represents the baseline target task performance of BERT without intermediate fine-tuning. \textbf{\textsc{TaskEmb} generally selects source tasks that yield positive transfer, and often selects the best source task.}
}
\label{figure:transfer}
\end{figure*}
\paragraph{In-class transfer:} The diagonal of each block  of Table~\ref{table:transfer} shows the results for in-class transfer, in which source tasks are from the same class as the target task. Across all three data regimes, most target tasks benefit from in-class transfer, and the average transfer gain is larger for CR and QA tasks than for SL tasks. Changing the data regimes significantly impacts the average transfer gain, which is lowest in the \smallsc{Full $\rightarrow$ Full} regime (+5.4\% average relative gain across all tasks) and highest in the \smallsc{Full $\rightarrow$ Limited} regime (+47.0\%). In general, tasks with fewer training examples benefit the most from transfer, such as RTE (+17.0 accuracy points) and CQ (+14.9 F1), and the best source tasks in the \smallsc{Full $\rightarrow$ Full} regime tend to be data-rich tasks such as MNLI, SNLI, and SQuAD-2 (Figure~\ref{figure:transfer}).\footnote{As in~\citet{JPhang18}, we find that intermediate fine-tuning reduces variance across random restarts (Appendix~\ref{appendix:b}).}

\paragraph{Out-of-class transfer:} %Having analyzed the effects of in-class transfer in different data regimes, 
We switch gears now to out-of-class transfer, in which the source task comes from a different class than the target task. The off-diagonal entries of each block of Table~\ref{table:transfer} summarize our results. In general, we observe that most tasks benefit from out-of-class transfer, although the magnitude of the transfer gains is lower than for in-class transfer, and that CR and QA tasks benefit more than SL tasks (similar to our in-class transfer results). While some of the results are intuitive (e.g., SQuAD is a good source task for QNLI, which is an entailment task built from QA pairs), others  are more difficult to explain (using part-of-speech tagging as a source task for DROP results in huge transfer gains in limited target regimes).

\paragraph{Large source datasets are not always best for data-constrained target tasks:} \citet{JPhang18} observe that source data size is a good heuristic to obtain positive transfer gain. In the \smallsc{Full $\rightarrow$ Limited} regime, we find to the contrary that the largest source datasets do not always result in the largest transfer gains. For CR tasks, MNLI/SNLI are the best sources for only four targets (three of which are entailment tasks), compared to seven in \smallsc{Full $\rightarrow$ Full}. STS-B, which is much smaller than MNLI and SNLI, is the best source for MRPC and QQP, while MRPC, an even smaller dataset, is the best source for STS-B. As STS-B, QQP, and MRPC are all sentence similarity and paraphrase tasks, this result suggests that the similarity between the source and target tasks matters more for data-constrained targets. We observe similar task similarity patterns for QA (the best source for WikiHop is the other multi-hop QA task, HotpotQA) and SL (POS-PTB is the best source for POS-EWT, the only other POS tagging task). However, the large SQuAD-2 dataset is almost always the best source within QA. Another important factor especially apparent in our QA tasks is domain similarity (e.g., SQuAD 
and several other datasets were all built from Wikipedia).

\paragraph{When does transfer work with data-constrained sources?}  We now turn to the \smallsc{Limited $\rightarrow$ Limited} regime, which eliminates the source data size confound. For CR, STS-B is the best source for six targets out of 11, including four entailment tasks (MNLI, QNLI, SNLI, SciTail), whereas MNLI/SNLI are the best sources for only two tasks (RTE, WNLI). This result suggests that source/target task similarity, which we found to be a factor for the \smallsc{Full $\rightarrow$ Limited}, is not the only important factor for effective transfer in data-constrained scenarios. We hypothesize that the complexity of the source task can also play a role: perhaps regression objectives (as used in STS-B) are more useful for transfer learning than classification objectives (MNLI/SNLI). Unknown factors may also play a role: in QA, SQuAD-2 is no longer the best source for any targets, while NewsQA is the best source for five tasks.

\section{Predicting task transferability}
\label{sec:taskemb}

The above analysis suggests that no single factor (e.g., data size, task and domain similarity, task complexity) is predictive of transfer gain across all of our settings. \textit{Given a novel target task, how can we identify the single source task that maximizes transfer gain?} One straightforward but extremely expensive approach is to enumerate every possible (source, target) task combination. Work on multi-task learning within NLP offers a more practical alternative by developing feature-based models to identify task and dataset characteristics that are predictive of task synergies~\cite{JBingel17}. Here, we take a different approach, inspired by recent computer vision methods~\citep{AAchille19}, by computing \emph{task embeddings} from layer-wise gradients of BERT. Our approach generally outperforms baseline methods that use the data size heuristic~\cite{JPhang18} and the gradients of the learning curve~\cite{JBingel17} in terms of selecting the most transferable source tasks across settings.

\subsection{Task embedding methods}
We develop two methods for computing task embeddings from BERT. The first, \textemb, is computed by pooling BERT's representations across an entire dataset, and as such captures properties of the text and domain. The second, \taskemb, relies on the correlation between the fine-tuning loss function and the parameters of BERT, and encodes more information about the type of knowledge and reasoning required to solve the task.

\paragraph{\textemb:} As our analysis indicates that domain similarity is a relevant factor for transfer, we first explore a simple method based on averaging BERT token-level representations of the inputs. Given a dataset $D$, we process each input sample $x^{i}$ through the pretrained BERT model without any finetuning and compute $\bvec{h}_x$, the average of final layer token-level representations. 
The final task embedding is the average of these pooled vectors over the entire dataset: $\sum_{x\in D} \dfrac{\bvec{h}_x}{|D|}$. This method captures linguistic properties of the input text $x$ and does not depend on the training labels $y$.

\paragraph{\taskemb:} Ideally, we want a way of capturing task similarity beyond just input properties represented by \textemb. Following the methodology of \ttv~\cite{AAchille19}, which develops task embeddings for meta-learning over vision tasks, we create representations of tasks derived from the Fisher information matrix (or simply \textit{Fisher}).  The Fisher captures the curvature of the loss surface (the sensitivity of the loss to small perturbations of model parameters), which intuitively tells us which of the model parameters are most useful for the task and thus provides a rich source of knowledge about the task itself.

To begin, we fine-tune BERT on the training dataset of a given task; the model without the final task-specific layer forms our \emph{feature extractor}. Next, we feed the entire training dataset into the model and compute the task embedding based on the Fisher of the feature extractor's parameters (weights) $\theta$, i.e., the expected covariance of the gradients of the log-likelihood with respect to $\theta$: \\
\begin{equation*}
F_{\theta} = 
\displaystyle \mathop{\mathbb{E}}_{x,y \sim P_\theta(x,y)}
 \nabla_{\theta} \log P_\theta(y|x) \nabla_{\theta} \log P_\theta(y|x)^{T}\text{.}   
\end{equation*}
In our experiments, we compute the \textit{empirical Fisher}, which uses the training labels instead of sampling from $P_\theta(x,y)$:\\
\scalebox{0.98}{\parbox{.5\linewidth}{%
\begin{align*}
F_{\theta} = \frac{1}{n}\sum\limits_{i=1}^{n} \left[ \nabla_{\theta} \log P_\theta(y^i|x^i) \nabla_{\theta} \log P_\theta(y^i|x^i)^{T}\right]\text{,}
\end{align*}
}}\\
and only consider the diagonal entries to reduce computational complexity. Additionally, we consider the Fisher $F_{\phi}$ with respect to the feature extractor's outputs (activations) $\phi$, which encodes useful features about the inputs to solve the task. The diagonal $F_{\phi}$ is averaged over the input tokens and over the entire dataset.\footnote{While Fisher matrices are theoretically more comparable when the feature extractor is fixed during fine-tuning, as done in \ttv, we find empirically that \taskemb\ computed from a fine-tuned task-specific BERT result in better correlations to task transferability in data-constrained scenarios. We leave further exploration of this phenomenon to future work.}

We explore task embeddings derived from the diagonal Fisher of different components of BERT, including the token embeddings, multi-head attention, feed-forward network, and the layer output, performing layer-wise averaging. Since our base model is BERT, this method may result in high-dimensional task embeddings (from 768-d to millions of dimensions). While one can optionally perform dimensionality reduction (e.g., through PCA), all of our experiments are conducted directly on the original task embeddings.
 
\subsection{Task embedding evaluation}
\label{sec:3.2}
We investigate whether a high similarity between two different task embeddings correlates with a high degree of transferability between those two tasks. Our evaluation centers around the meta-task of selecting the best source task for a given target task. Specifically, given a target task, we rank all the other source tasks in our library in descending order by the cosine similarity\footnote{We leave the exploration of asymmetric similarity metrics to future work.} between their task embeddings and the target task’s embedding. This ranking is evaluated using two metrics: (1) the average rank $\rho$ of the source task with the highest absolute transfer gain from Section~\ref{sec:transfer}'s experiments, and (2) the Normalized Discounted Cumulative Gain~\cite[NDCG;][]{KJarvelin02}, a common information retrieval measure that evaluates the quality of the entire ranking, not just the rank of the best source task.\footnote{We use NDCG instead of Spearman correlation, as the latter penalizes top-ranked and bottom-ranked mismatches with the same weight.} The NDCG at position $p$ is defined as:
\scalebox{0.90}{\parbox{.5\linewidth}{%
\begin{align*}
\text{NDCG}_{p} = \dfrac{\text{DCG}_{p}(R_{pred})}{\text{DCG}_{p}(R_{true})}
\end{align*}
}}, where $R_{pred}, R_{true}$ are the predicted and gold rankings of the source tasks, respectively; and 
\scalebox{0.90}{\parbox{.5\linewidth}{%
\begin{align*}
\text{DCG}_{p}(R) = \sum\limits_{i=1}^{p} \dfrac{2^{rel_i} - 1}{\log_{2}(i+1)}
\end{align*}
}}, where $rel_{i}$ is the relevance (target performance) of the source task with rank $i$ in the evaluated ranking $R$.\footnote{In our experiments, we set $p$ to the number of source tasks in each setting.} An \text{NDCG} of 100\% indicates a perfect ranking.

\paragraph{Aggregating similarity signals from embedding spaces: } 
For our \taskemb\ approach, we aggregate rankings from all of the different components of BERT rather than evaluate each component-specific ranking separately.\footnote{We observe that rankings derived from certain components are more useful than others (e.g., token embeddings are crucial for classification), but aggregating across all components generally outperforms individual ones.}
We expect that task embeddings derived from different components might contain complementary information about the task, which motivates this decision.
Concretely, given a target task $t$, assume that $r_{1:c}$ are the rank scores assigned to a source task $s$ by $c$ different components of BERT. Then, the aggregated score is computed according to the reciprocal rank fusion algorithm~\cite{GCormack09}: 
%$\textsc{rrf}(s) = \sum_{i=1}^c{1/(60 + r_i)}$. 
\scalebox{0.90}{\parbox{.5\linewidth}{
\begin{align*}
\text{RRF}(s) = \sum\limits_{i=1}^{c}\dfrac{1}{60 + r_i}
\end{align*}
}}. We also use this approach to aggregate rankings from \textemb\ and \taskemb, which results in \textsc{Text + Task}.

\subsection{Baseline methods}
\paragraph{\datasize: }
To measure the effect of data size, we compare rankings derived from \textemb\ and \taskemb\ to \datasize, a heuristic baseline that ranks all source tasks by the number of training examples. 
\paragraph{\curvegrad: } We also consider \curvegrad, a baseline that uses the gradients of the loss curve of BERT for each task. ~\citet{JBingel17} find such learning curve features to be good predictors of gains from multi-task learning. %for sequence labeling tasks. 
They suggest that multi-task learning is more likely to work when the main tasks quickly plateau (small negative gradients) while the auxiliary tasks continue to improve (large negative gradients). Following the setup in~\citet{JBingel17}, we fine-tune BERT on each source task for a fixed number of steps (i.e., 10,000) and compute the gradients of the loss curve at 10, 20, 30, 50 and 70 percent of the fine-tuning process. Given a target task, we rank all the source tasks in descending order by the gradients and aggregate the rankings using the reciprocal rank fusion algorithm.

\subsection{Source task selection experiments}
\begin{table*}[t!]
\centering
\begin{adjustbox}{max width=\textwidth}
\begin{tabular}{p{2cm} rrrr| rrrr|  rrrr}
\toprule
& \multicolumn{4}{c}{\smallsc{Full $\rightarrow$ Full}} &
\multicolumn{4}{c}{\smallsc{Full $\rightarrow$ Limited}} &
\multicolumn{4}{c}{\smallsc{Limited $\rightarrow$ Limited}}\\
\cmidrule(l){2-5} \cmidrule(l){6-9} \cmidrule(l){10-13}
& \multicolumn{2}{c}{\emph{in-class (10)}} & \multicolumn{2}{c}{\emph{all-class (32)}} & \multicolumn{2}{c}{\emph{in-class (10)}} & \multicolumn{2}{c}{\emph{all-class (32)}} & \multicolumn{2}{c}{\emph{in-class (10)}} & \multicolumn{2}{c}{\emph{all-class (32)}}\\
\cmidrule(lr){2-3} \cmidrule(lr){4-5} 
\cmidrule(lr){6-7} \cmidrule(lr){8-9} 
\cmidrule(lr){10-11} \cmidrule(lr){12-13}  
\textbf{Method} & \multicolumn{1}{l}{$\rho$} & \multicolumn{1}{l}{NDCG} & \multicolumn{1}{l}{$\rho$} & \multicolumn{1}{l}{NDCG} &\multicolumn{1}{l}{$\rho$} & \multicolumn{1}{l}{NDCG} &\multicolumn{1}{l}{$\rho$} & \multicolumn{1}{l}{NDCG} &\multicolumn{1}{l}{$\rho$} & \multicolumn{1}{l}{NDCG} & \multicolumn{1}{l}{$\rho$} & \multicolumn{1}{l}{NDCG} \\
\midrule 
\multicolumn{5}{l}{\emph{classification / regression}}\\[0.5ex] 
\datasize &3.6	&80.4	&8.5	&74.7	
&3.8	&62.9	&9.8	&54.6 &
\multicolumn{1}{c}{-}&\multicolumn{1}{c}{-}&\multicolumn{1}{c}{-}&\multicolumn{1}{c}{-}\\
\curvegrad 	&5.5	&68.6	&17.8	&64.9	&6.4	&45.2	&18.8	&35.0	&5.9	&50.8	&13.3	&42.4\\
\textemb	&5.2	&76.4	&13.1	&71.3
&3.5	&60.3	&8.6	&52.4
&4.8	&61.4	&13.2	&43.9 \\
\taskemb	&2.8	&82.3	&6.2	&76.7
&3.4	&68.2	&\textbf{8.2}	&60.9
&\textbf{4.2}	&62.6	&11.6	&\textbf{44.8} \\
\textsc{Text+Task}	&\textbf{2.6}	&\textbf{83.3}	&\textbf{5.6}	&\textbf{78.0}
&\textbf{3.3}	&\textbf{69.5}	&\textbf{8.2}	&\textbf{62.0}
&\textbf{4.2}	&\textbf{62.7}	&\textbf{11.4}	&\textbf{44.8}\\
\midrule 
\multicolumn{5}{l}{\emph{question answering}}\\[0.5ex] 
\datasize	&\textbf{3.2}	&84.4	&13.8	&63.5
&2.3	&77.0	&13.6	&40.2
&\multicolumn{1}{c}{-}&\multicolumn{1}{c}{-}&\multicolumn{1}{c}{-}&\multicolumn{1}{c}{-}\\	
\curvegrad 	&8.3	&64.8	&15.7	&55.0	&8.2	&49.1	&16.7	&32.8	&6.8	&53.4	&15.3	&40.1\\
\textemb	&4.1	&81.1	&6.8	&79.7
&2.7	&77.6	&4.1	&77.0
&4.1	&65.6	&7.6	&66.5\\
\taskemb	&\textbf{3.2}	&84.5	&6.5	&81.6
&2.5	&78.0	&4.0	&79.0
&\textbf{3.6}	&\textbf{67.1}	&7.5	&68.5\\
\textsc{Text+Task}	&\textbf{3.2}	&\textbf{85.9}	&\textbf{5.4}	&\textbf{82.5}
&\textbf{2.2}	&\textbf{81.2}	&\textbf{3.6}	&\textbf{82.0}
&\textbf{3.6}	&66.5	&\textbf{7.0}	&\textbf{69.6}\\
\midrule
\multicolumn{5}{l}{\emph{sequence labeling}}\\[0.5ex] 
\datasize	&7.9	&90.5	&19.2	&91.6
&4.3	&63.2	&20.3	&34.0
&\multicolumn{1}{c}{-}&\multicolumn{1}{c}{-}&\multicolumn{1}{c}{-}&\multicolumn{1}{c}{-}\\
\curvegrad 	&5.6    &92.6   &14.6   &92.8   &8.0   &40.7   &17.9   &30.8   &7.0   &53.2   &18.6   &40.8\\
\textemb	&3.7	&95.0	&10.4	&\textbf{95.3}
&3.9	&65.1	&8.5	&61.1
&5.0	&67.2	&10.1	&63.8 \\
\taskemb	&3.4	&95.7	&\textbf{9.6}	&95.2
&\textbf{2.7}	&\textbf{80.5}	&4.4	&76.3
&\textbf{2.5}	&82.1	&5.5	&\textbf{76.9} \\
\textsc{Text+Task}	&\textbf{3.3}	&\textbf{96.0}	&\textbf{9.6}	&95.2
&\textbf{2.7}	&80.3	&\textbf{4.2}	&\textbf{78.4}
&\textbf{2.5}	&\textbf{82.5}	&\textbf{5.3}	&\textbf{76.9} \\
\bottomrule
\end{tabular}
\end{adjustbox}
\caption{To evaluate our embedding methods, we measure the average rank ($\rho$) that they assign to the best source task (i.e., the one that results in the largest transfer gain) across target tasks, as well as the average NDCG measure of the overall ranking's quality. In parentheses, we show the number of source tasks in each setting. Combining the complementary signals in \taskemb\ and \textemb\ consistently decreases $\rho$ (lower is better) and increases NDCG across all settings, and both methods in isolation generally perform better than the baseline methods.}
\label{table:taskemb}
\vspace{-1.5mm}
\end{table*}
The average performance of selecting the best source task across target tasks using different methods is shown in Table~\ref{table:taskemb}.\footnote{In the \smallsc{Limited} settings, we report the mean results across random restarts.} Here, we provide an overview and analysis of these results.

\paragraph{Baselines:}
\datasize\ is a good heuristic when the full source training data is available, but it struggles in all out-of-class transfer scenarios as well as on SL tasks, for which most datasets contain roughly the same number of examples (Table~\ref{table:datasets}).\footnote{All methods obtain a higher NDCG score on SL tasks in the \smallsc{Full $\rightarrow$ Full} regime because there is little difference in target task performance between source tasks here (see Figure~\ref{figure:transfer}), and thus the rankings are not penalized heavily.} \curvegrad\ lags far behind \datasize\ in most cases, though its performance is better on SL tasks in the \smallsc{Full $\rightarrow$ Full} regime. This indicates that \curvegrad\ cannot reliably predict the most transferable source tasks in our transfer scenarios.

\paragraph{\textemb\ and \taskemb\ improve transferability prediction: }

Table~\ref{table:taskemb} shows that \textemb\ performs better than \datasize\ on average, especially within the limited data regimes. Interestingly, \textemb\ underperforms  significantly on CR tasks compared to QA and SL. We theorize that this effect is partly due to the relative homogeneity of the QA and SL datasets (i.e., many QA datasets use Wikipedia while many SL tasks are extracted from the Penn Treebank) compared to the more diverse CR datasets. If \textemb\ captures mainly domain similarity, then it may struggle when that is not a relevant transfer factor.

\taskemb\ can substantially boost the quality of the rankings, frequently outperforming the other methods across different classes of problems, data regimes, and transfer scenarios. These results demonstrate that the task similarity between the computed embeddings is a robust predictor of effective transfer.  The ensemble of \textsc{Text + Task} results in further slight improvements, but the small magnitude of these gains suggests that \taskemb\ partially encodes domain similarity. For \smallsc{Limited $\rightarrow$ Limited}, where the \datasize\ heuristic does not apply, \taskemb\ still performs strongly, although not as well as in the full source data regimes. Figure~\ref{figure:transfer} shows that \taskemb\ usually selects the best or near the best available source task for a given target task across data regimes.

\paragraph{Understanding the task embedding spaces:}
\begin{figure}[ht!]
\centering
\includegraphics[width=0.48\textwidth]{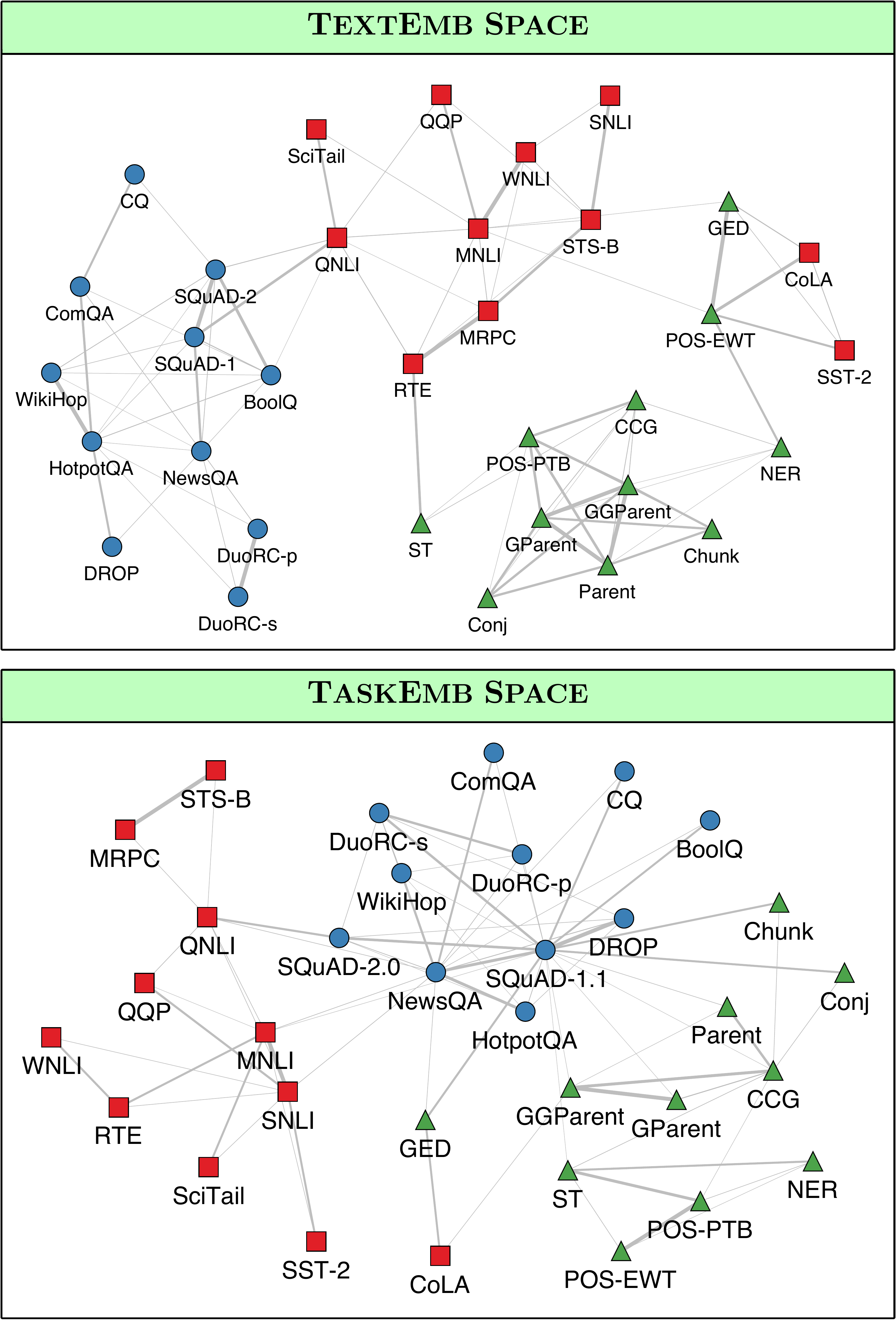}
\caption{A 2D visualization of the task spaces of \textsc{TextEmb} and \textsc{TaskEmb}. \textemb\ captures domain similarity (e.g., the Penn Treebank SL tasks are highly interconnected), while \taskemb\ focuses more on task similarity (the two part-of-speech tagging tasks are interconnected despite their domain dissimilarity).}
\label{figure:space}
\vspace*{-2mm}
\end{figure} What kind of information is
encoded by \taskemb\ and \textemb? Figure~\ref{figure:space} visualizes the different task spaces in the \smallsc{Full $\rightarrow$ Full} regime using the Fruchterman-Reingold force-directed placement algorithm~\cite{TFruchterman91}.\footnote{An alternative to dimensionality reduction algorithms for better preservation of the data's topology; see Appendix~\ref{appendix:a2}.}

The task space of \textemb\ (Figure~\ref{figure:space}, top) shows that datasets with similar sources are near one another: in QA, tasks built from web snippets are closely linked (CQ and ComQA), while in SL, tasks extracted from Penn Treebank are clustered together (CCG, POS-PTB, Parent, GParent, GGParent, Chunk, and Conj). Additionally, the SQuAD datasets are strongly linked to QNLI, which was created by converting SQuAD questions. \taskemb\  captures domain information to some extent (Figure~\ref{figure:space}, bottom), but  it also encodes task similarity: for example, POS-PTB is  closer to POS-EWT, another part-of-speech tagging task that uses a different data source. Neither method captures some unintuitive cases in low-data regimes, such as STS-B's high transferability to CR target tasks, or that DROP benefits most from SL tasks in low-data regimes (see Tables~\ref{tbla4b},~\ref{tbla4c},~\ref{tbla10b}, and~\ref{tbla10c} in Appendix~\ref{appendix:b}).
Our methods clearly do not capture all of the factors that influence task transferability, which motivates the future development of more sophisticated task embedding methods.
\section{Related Work}
\label{section:related_work}

We build on existing work in exploring and predicting transferability across tasks.

\paragraph{Transferability between NLP tasks:} Sharing knowledge across different tasks, as in multi-task/transfer learning, often improves over standard single-task learning~\cite{SRuder17}. Within multi-task learning, several works~\citep[e.g.,][]{TLuong16,XLiu19,Raffel20}
combine multiple tasks for better regularization and transfer. More related to our work, \citet{JPhang18} explore intermediate fine-tuning and find that transferring from data-rich source tasks boosts target task performance for text classification, while \citet{NLiu19} observe transfer gains between related sequence labeling tasks. Expanding from single to multi-source transfer, \citet{ATalmor19} show that pretraining on multiple datasets improves generalization on QA tasks. Nevertheless, exploiting synergies between tasks remains difficult, with many combinations of tasks negatively impacting downstream performance~\cite{JBingel17,BMcCann18,AWang19c}, and the factors that determine successful transfer still remain murky. Concurrent work indicates that intermediate tasks that require high-level inference and reasoning abilities tend to work best~\citep{YPruksachatkun20}.

\paragraph{Identifying beneficial task relationships:}  To predict transferable tasks, some methods~\citep{MAlonso17,JBingel17} rely on features derived from dataset characteristics and learning curves. However, manually designing such features is time-consuming and may not generalize well across classes of problems~\cite{EKerinec18}. Recent work on \emph{task embeddings} in computer vision offers a more principled way to encode tasks for meta-learning~\citep{AZamir18,AAchille19,XYan20}. Taskonomy~\cite{AZamir18} models the underlying structure among tasks to reduce the need for supervision, while Task2Vec~\cite{AAchille19} uses a frozen feature extractor pretrained on ImageNet to represent tasks in a topological space (analogous to our approach's reliance on BERT). Finally, recent work in NLP augments a generative model with an embedding space for modeling latent skills~\cite{KCao20}.
\section{Conclusion}

We conduct a large-scale empirical study of the transferability between 33 NLP tasks across three broad classes of problems. We show that the benefits of transfer learning are more pronounced than previously thought, especially when target training data is limited, and we develop methods that learn vector representations of tasks that can be used to reason about the relationships between them. These \emph{task embeddings} allow us to predict source tasks that will likely improve target task performance. Our analysis suggests that data size, the similarity between the source and target tasks and domains, and task complexity are crucial for effective transfer, particularly in data-constrained regimes. 
\section*{Acknowledgments}

We thank Yoshua Bengio and researchers at Microsoft Research Montreal for valuable feedback on this project. We also thank the anonymous reviewers, Kalpesh Krishna, Nader Akoury, Shiv Shankar, and the rest of the UMass NLP group for their helpful comments. We are grateful to Alon Talmor and Nelson Liu for sharing the QA and SL datasets. Finally, we thank Peter Potash for additional experimentation efforts. Vu and Iyyer were supported by an Intuit AI Award for this project.
\clearpage
\newpage
%\bibliography{emnlp2020}
%\bibliographystyle{acl_natbib}

\clearpage
\newpage
\appendix

\section*{Appendices}
\section{Additional details for experimental setup}
\subsection{Tasks \& datasets} 
\label{appendix:a1}
\begin{table*}[t!]
\centering
\begin{adjustbox}{max width=\textwidth}
\begin{tabular}{llll}
\toprule
\textbf{Task} & \textbf{$\mid$ Train $\mid$} & \textbf{Task type} & \textbf{Domain} \\ [0.5ex] 
\multicolumn{4}{l}{\emph{text classification/regression (CR)}} \\ [0.5ex] 
SNLI~\cite{SBowman15} & 570K & NLI & misc.\\
MNLI~\cite{AWilliams18} & 393K & NLI & misc. % image captions
\\
QQP~\cite{Quora17} & 364K & paraphrase identification & social QA \\
QNLI~\cite{AWang19a} & 105K & QA-NLI & Wikipedia \\
SST-2~\cite{RSocher13} &  67K & sentiment analysis & movie reviews\\
SciTail~\cite{TKhot18} & 27K & NLI & science QA\\
CoLA~\cite{AWarstadt19} &  8.5K & grammatical acceptability & misc.%books, journal articles on linguistic theory
\\
STS-B~\cite{DCer17} & 7K & semantic similarity & misc. \\
MRPC~\cite{WDolan05} & 3.7K & paraphrase identification & news \\
RTE~\cite[][et seq.]{IDagan06} & 2.5K & NLI & news, Wikipedia \\
WNLI~\cite{HLevesque11} & 634 & coreference NLI & fiction books \\
\midrule 
\multicolumn{4}{l}{\emph{question answering (QA)}} \\ [0.5ex] 
SQuAD-2~\cite{PRajpurkar18} & 162K & QA & Wikipedia, crowd \\
NewsQA~\cite{ATrischler17} & 120K & QA & news, crowd \\
HotpotQA~\cite{ZYang18} & 113K & multi-hop QA & Wikipedia, crowd \\
SQuAD-1~\cite{PRajpurkar16} & 108K & QA & Wikipedia, crowd \\
DuoRC-p~\cite{ASaha18} & 100K & paraphrased QA & Wikipedia/IMDB, crowd\\
DuoRC-s~\cite{ASaha18} & 86K & paraphrased QA & Wikipedia/IMDB, crowd\\
DROP~\cite{DDua19} & 77K & multi-hop quantitative reasoning & Wikipedia, crowd \\
WikiHop~\cite{JWelbl18} & 51K & multi-hop QA & Wikipedia, KB \\
BoolQ~\cite{CClark19} & 16K & natural yes/no QA & Wikipedia, web queries \\
ComQA~\cite{AAbujabal19} & 11K & factoid QA w/ paraphrases & snippets, WikiAnswers \\
CQ~\cite{JBao16} & 2K & knowledge-based QA & snippets, web queries/KB \\
 \midrule
 \multicolumn{4}{l}{\emph{sequence labeling (SL)}} \\ [0.5ex] 
ST~\cite{JBjerva16} & 43K & semantic tagging & Groningen Meaning Bank\\
CCG~\cite{JHockenmaier07} & 40K & CCG supertagging & Penn Treebank \\
Parent~\cite{NLiu19} & 40K & syntactic tagging & Penn Treebank\\
GParent~\cite{NLiu19} & 40K & syntactic tagging & Penn Treebank\\
GGParent~\cite{NLiu19} & 40K & syntactic tagging & Penn Treebank\\
POS-PTB~\cite{MMarcus93} & 38K & part-of-speech tagging & Penn Treebank \\
GED~\cite{HYannakoudakis11} & 29K & grammatical error detection & misc. \\
NER~\cite{TKSang03} & 14K & named entity recognition & news\\
POS-EWT~\cite{NSilveira14} & 13K & part-of-speech tagging & Web Treebank \\
Conj~\cite{JFicler16} & 13K & conjunct identification & Penn Treebank\\
Chunk~\cite{TKSang00} & 9K & syntactic chunking & Penn Treebank\\
\bottomrule
\end{tabular}
\end{adjustbox}
\caption{Datasets used in our experiments and their characteristics, grouped by task class and sorted by training dataset size.}
\label{table:dataset_characteristics}
%\vspace*{-2mm}
\end{table*}
In this work, we experiment with 33 datasets across three broad classes of problems (text classification/regression, question answering, and sequence labeling). Below, we briefly describe the datasets, and summarize their characteristics in Table~\ref{table:dataset_characteristics}.
\paragraph{Text classification/regression (eleven tasks):} We use the nine GLUE datasets~\cite{AWang19a}, including grammatical acceptability judgments~\cite[\textbf{CoLA};][]{AWarstadt19}; sentiment analysis~\cite[\textbf{SST-2};][]{RSocher13}; paraphrase identification~\cite[\textbf{MRPC};][]{WDolan05}; semantic similarity with STS-Benchmark~\cite[\textbf{STS-B};][]{DCer17} and Quora Question Pairs\footnote{\href{https://data.quora.com/First-Quora-Dataset-Release-Question-Pairs}{https://data.quora.com/First-Quora-Dataset-Release-Question-Pairs}} (\textbf{QQP}); natural language inference (NLI) with Multi-Genre NLI~\cite[\textbf{MNLI};][]{AWilliams18}, SQuAD~\cite{PRajpurkar16}
converted into Question-answering NLI~\cite[\textbf{QNLI};][]{AWang19a}, Recognizing Textual Entailment 1,2,3,5~\cite[\textbf{RTE};][et seq.]{IDagan06}, and the Winograd Schema Challenge~\cite{HLevesque11} recast as Winograd NLI (\textbf{WNLI}). Additionally, we include the Stanford NLI dataset~\cite[\textbf{SNLI};][]{SBowman15} and the science QA dataset~\cite{TKhot18} converted into NLI (\textbf{SciTail}). We report F1 scores for QQP and MRPC, Spearman correlations for STS-B, and accuracy scores for the other tasks. For MNLI, we report the average score on the ``matched" and ``mismatched" development sets.
\paragraph{Question answering (eleven tasks):} We use eleven QA datasets from the MultiQA~\cite{ATalmor19} repository\footnote{\href{https://github.com/alontalmor/MultiQA}{https://github.com/alontalmor/MultiQA}}, including the Stanford Question Answering datasets \textbf{SQuAD-1} and \textbf{SQuAD-2}~\cite{PRajpurkar16,PRajpurkar18}; \textbf{NewsQA}~\cite{ATrischler17}; \textbf{HotpotQA}~\cite{ZYang18} -- the version where the context includes 10 paragraphs retrieved by an information retrieval system; Natural Yes/No Questions dataset~\cite[\textbf{BoolQ};][]{CClark19}; Discrete Reasoning Over Paragraphs dataset~\cite[\textbf{DROP};][]{DDua19} -- we only use the extractive examples in the original dataset but evaluate on the entire development set, following~\citet{ATalmor19}; \textbf{WikiHop}~\cite{JWelbl18}; DuoRC Self (\textbf{DuoRC-s}) and DuoRC Paraphrase (\textbf{DuoRC-p}) datasets~\cite{ASaha18} where the questions are taken from either the same version or a different version of the document from which the questions were asked, respectively; ComplexQuestions~\cite[\textbf{CQ};][]{JBao16,ATalmor17}; and \textbf{ComQA}~\cite{AAbujabal19} -- contexts are not provided but the questions are augmented with web snippets retrieved from Google search engine~\cite{ATalmor19}. We report F1 scores for all QA tasks.
\paragraph{Sequence labeling (eleven tasks):} We experiment with eleven sequence labeling tasks used by~\citet{NLiu19}, including CCG supertagging with CCGbank~\cite[\textbf{CCG};][]{JHockenmaier07}; part-of-speech tagging with the Penn Treebank~\cite[\textbf{POS-PTB};][]{MMarcus93} and the Universal Dependencies English Web Treebank~\cite[\textbf{POS-EWT};][]{NSilveira14}; syntactic constituency ancestor tagging, i.e., predicting the constituent label of the parent (\textbf{Parent}), grandparent (\textbf{GParent}), and great-grandparent (\textbf{GGParent}) of each word in the PTB phrase-structure tree; semantic tagging task~\cite[\textbf{ST};][]{JBjerva16,LAbzianidze17}; syntactic chunking with the CoNLL 2000 shared task dataset~\cite[\textbf{Chunk};][]{TKSang00}; named entity recognition with the CoNLL 2003 shared task dataset~\cite[\textbf{NER};][]{TKSang03}; grammatical error detection with the First Certificate in English dataset~\cite[\textbf{GED};][]{HYannakoudakis11,MRei16}; and conjunct identification, i.e., identifying the tokens that comprise the conjuncts in a coordination construction, with the coordination annotated PTB dataset~\cite[\textbf{Conj};][]{JFicler16}. We report F1 scores for all SL tasks.
\newpage
\subsection{Fruchterman-Reingold force-directed placement algorithm}
\label{appendix:a2}
The Fruchterman-Reingold force-directed placement algorithm~\cite{TFruchterman91} simulates a space of nodes (in our setup, tasks) as a system of atomic particles/celestial bodies, exerting attractive forces on one another. In our setup, the algorithm resembles molecular/planetary simulations: the transferability between tasks specify the forces that are used to place the tasks towards each other in order to minimize the energy of the system. The force between a pair of tasks $(t_1, t_2)$ is defined as: \scalebox{0.90}{\parbox{.5\linewidth}{%
\begin{align*}
f(t_1,t_2) = \dfrac{1}{r_{\rightarrow t_2}(t_1)} + \dfrac{1}{r_{\rightarrow t_1}(t_2)}
\end{align*}
}}, where $r_{\rightarrow t}(s)$ is the rank of the source task $s$ in the list of source tasks to transfer to the target task $t$.
\clearpage
\newpage
\section{Full results for fine-tuning and transfer learning across tasks}
\label{appendix:b}
For both fine-tuning and transfer learning, we use the same architecture across tasks, apart from the task-specific output layer. The feature extractor, i.e., BERT, is pretrained while the task-specific output layer is randomly initialized for each task. All the parameters are fine-tuned end-to-end. An alternative approach is to keep the feature extractor frozen during fine-tuning. We find that fine-tuning the whole model for a given task leads to better performance in most cases, except for WNLI and DROP, possibly because of their adversarial nature (see Tables~\ref{tbla1}, ~\ref{tbla2}, and~\ref{tbla3}). % Finetuning procedure
In our experiments, we follow the fine-tuning recipe of~\cite{JDevlin19}, i.e., only fine-tuning for a fixed number of $t$ epochs for each class of problems. We develop our infrastructure using the HuggingFace's Transformers~\cite{TWolf19} and its
recommended hyperparameters for each class.

We show the full results for fine-tuning and transfer learning across tasks from Table~\ref{tbla1} to Table~\ref{tbla12c}. Below, we describe the setting for these tables in more detail:

In Tables~\ref{tbla1}, ~\ref{tbla2}, and ~\ref{tbla3}, we report the results of fine-tuning BERT (without any intermediate fine-tuning) on the 33 NLP tasks studied in this work. We perform experiments in two data regimes: \smallsc{Full} and \smallsc{Limited}. In the \smallsc{Full} regime, all training data for the associated task is used while in the \smallsc{Limited} setting, we artificially limit the amount of training data by randomly selecting 1K training examples without replacement, following~\citet{JPhang18}. For each experiment in the \smallsc{Limited} regime, we perform 20 random restarts (1K examples are resampled for each restart) and report the mean and standard deviation. We show the results after each training epoch $t$.

For our transfer experiments, we consider every possible pair of (source, target) tasks within and across classes of problems in the three data regimes described in~\ref{sec:data_regimes}, which results in 3267 combinations of tasks and data regimes. We follow the transfer recipe of~\citet{JPhang18} by first fine-tuning BERT on the source task (intermediate fine-tuning) before fine-tuning on the target task. For both stages, we only perform training for a fixed number $t$ of epochs, following previous work~\cite{JDevlin19,JPhang18}. For each task, we use the same value of $t$ as in our fine-tuning experiments. \\

From Table~\ref{tbla4a} to Table~\ref{tbla6c}, we show our in-class transfer results for each combination of (source, target) tasks, in which source tasks come from the same class as the target task. In each table, rows denote source tasks while columns denote target tasks. Each cell represents the target task performance of the transferred model from the associated source task to the associated target task. The orange-colored cells along the diagonal indicate the results of fine-tuning BERT on target tasks without any intermediate fine-tuning. Positive transfers are shown in blue and the best results are highlighted in bold (blue). For transfer results in the \smallsc{Limited} setting, we report the mean and standard deviation across 20 random restarts.

Finally, from Table~\ref{tbla7a} to Table~\ref{tbla12c}, we present our out-of-class transfer results, in which source tasks come from a different class than the target task. In each table, results are shown in a similar way as above, except that the orange-colored row Baseline shows the results of fine-tuning BERT on target tasks without any intermediate fine-tuning.

\clearpage
%\aclfinalcopy
\newpage 
\begin{table*}[ht]
\centering
\begin{adjustbox}{max width=\textwidth}
\begin{tabular}{ l  l l l | l l l | l l l}
  \toprule
\multirow{3}{*}{\textbf{Task}} %& \multicolumn{9}{c }{\textbf{Data regime}}\\ \cline{2-10}
& \multicolumn{6}{c}{\textsc{Full}} & \multicolumn{3}{c}{\textsc{Limited}} \\ \cmidrule(lr){2-7} \cmidrule(lr){8-10}  
& \multicolumn{3}{c}{\emph{frozen BERT}} &  \multicolumn{3}{c}{\emph{unfrozen BERT}} & \multicolumn{3}{c}{\emph{unfrozen BERT}} \\ 
\cmidrule(lr){2-4} \cmidrule(lr){5-7}  
& \multicolumn{1}{c}{t = 1} & \multicolumn{1}{c}{t = 2} & \multicolumn{1}{c}{t = 3} & \multicolumn{1}{c}{t = 1} & \multicolumn{1}{c}{t = 2} & \multicolumn{1}{c}{t = 3} & \multicolumn{1}{c}{t = 1} & \multicolumn{1}{c}{t = 2} & \multicolumn{1}{c}{t = 3}  \\
\midrule
CoLA & 0.0 & 0.0 & 0.0 & 48.1 & 51.3 & 51.0 & 1.0 $\pm$ 2.3 & 4.0 $\pm$ 7.4 & 4.7 $\pm$ 8.2\\
SST-2 & 51.0 & 51.5 & 51.9 & 91.4 & 92.1 & 91.9 & 61.5 $\pm$ 7.9 & 74.3 $\pm$ 8.2 & 77.5 $\pm$ 6.3\\
MRPC & 81.2 & 81.2 & 81.2 & 81.2 & 82.4 & 84.0 & 70.4 $\pm$ 26.2 & 81.8 $\pm$ 0.6 & 81.9 $\pm$ 0.7\\
STS-B & 68.0 & 68.3 & 68.4 & 76.7 & 85.4 & 85.9 & 3.6 $\pm$ 9.5 & 22.8 $\pm$ 10.5 & 29.9 $\pm$ 10.5\\
QQP & 0.2 & 13.9 & 16.9 & 86.0 & 87.0 & 87.3 & 9.5 $\pm$ 15.5 & 12.1 $\pm$ 15.9 & 25.7 $\pm$ 25.1\\
MNLI & 40.9 & 40.2 & 40.8 & 83.1 & 84.3 & 84.2 & 33.7 $\pm$ 3.1 & 37.5 $\pm$ 3.4 & 38.7 $\pm$ 3.2\\
QNLI & 65.9 & 66.0 & 66.0 & 90.3 & 91.3 & 91.4 & 58.0 $\pm$ 9.4 & 61.0 $\pm$ 9.9 & 62.4 $\pm$ 9.5\\
RTE & 53.8 & 53.1 & 51.3 & 56.0 & 58.1 & 60.6 & 50.7 $\pm$ 3.8 & 54.6 $\pm$ 3.4 & 54.7 $\pm$ 3.2\\
WNLI & 56.3 & 56.3 & 56.3 & 52.1 & 46.5 & 45.1 & 47.9 $\pm$ 5.6 & 45.6 $\pm$ 6.0 & 44.4 $\pm$ 6.3\\
SNLI & 42.2 & 43.4 & 44.9 & 90.3 & 90.8 & 90.7 & 40.2 $\pm$ 4.5 & 45.1 $\pm$ 4.9 & 46.7 $\pm$ 4.5\\
SciTail & 49.6 & 49.6 & 49.6 & 92.3 & 93.7 & 93.9 & 52.5 $\pm$ 6.3 & 60.1 $\pm$ 12.5 & 64.1 $\pm$ 13.6\\
\bottomrule
 \end{tabular}
 \end{adjustbox}
  \caption{Fine-tuning results for classification/regression tasks.}
  \label{tbla1}
\end{table*}
\clearpage
\newpage

\begin{table*}[t!]
\centering
\begin{adjustbox}{max width=\textwidth}
 \begin{tabular}{ l  l l l | l l l | l l l}
  \toprule
\multirow{3}{*}{\textbf{Task}} %& \multicolumn{9}{c }{\textbf{Data regime}}\\ \cline{2-10}
& \multicolumn{6}{c}{\textsc{Full}} & \multicolumn{3}{c}{\textsc{Limited}} \\ \cmidrule(lr){2-7} \cmidrule(lr){8-10}  
& \multicolumn{3}{c}{\emph{frozen BERT}} &  \multicolumn{3}{c}{\emph{unfrozen BERT}} & \multicolumn{3}{c}{\emph{unfrozen BERT}} \\ 
\cmidrule(lr){2-4} \cmidrule(lr){5-7}  
& \multicolumn{1}{c}{t = 1} & \multicolumn{1}{c}{t = 2} & \multicolumn{1}{c}{t = 3} & \multicolumn{1}{c}{t = 1} & \multicolumn{1}{c}{t = 2} & \multicolumn{1}{c}{t = 3} & \multicolumn{1}{c}{t = 1} & \multicolumn{1}{c}{t = 2} & \multicolumn{1}{c}{t = 3}  \\
\midrule
SQuAD-1 & 10.6 & 12.1 & 13.0 & 86.8 & 87.7 & 87.9 & 12.5 $\pm$ 1.0 & 20.8 $\pm$ 4.6 & 26.8 $\pm$ 6.0\\
SQuAD-2 & 49.8 & 49.8 & 49.8 & 68.4 & 70.4 & 71.9 & 50.0 $\pm$ 0.1 & 50.0 $\pm$ 0.1 & 50.1 $\pm$ 0.1\\
NewsQA & 9.4 & 10.4 & 10.6 & 64.7 & 64.9 & 64.1 & 15.6 $\pm$ 3.4 & 26.5 $\pm$ 4.7 & 28.8 $\pm$ 4.9\\
HotpotQA & 5.9 & 6.8 & 7.0 & 66.1 & 68.2 & 67.9 & 12.8 $\pm$ 2.4 & 21.6 $\pm$ 3.9 & 23.3 $\pm$ 4.0\\
BoolQ & 62.1 & 62.2 & 62.2 & 62.2 & 66.4 & 65.7 & 62.2 $\pm$ 0.0 & 62.2 $\pm$ 0.1 & 62.2 $\pm$ 0.0\\
DROP & 42.9 & 51.7 & 54.1 & 22.4 & 21.5 & 22.4 & 6.8 $\pm$ 4.4 & 13.5 $\pm$ 10.0 & 19.4 $\pm$ 11.8\\
WikiHop & 10.1 & 11.4 & 11.6 & 60.0 & 62.3 & 62.8 & 18.3 $\pm$ 4.0 & 24.8 $\pm$ 4.9 & 25.5 $\pm$ 4.7\\
DuoRC-p & 42.1 & 42.1 & 42.1 & 50.3 & 50.3 & 50.6 & 42.1 $\pm$ 0.0 & 42.2 $\pm$ 0.2 & 41.6 $\pm$ 1.1\\
DuoRC-s & 4.6 & 5.6 & 5.8 & 66.2 & 64.4 & 63.3 & 22.2 $\pm$ 11.0 & 37.5 $\pm$ 3.5 & 38.9 $\pm$ 3.3\\
CQ & 15.4 & 15.4 & 15.9 & 26.3 & 25.0 & 30.5 & 28.0 $\pm$ 3.3 & 29.6 $\pm$ 2.1 & 30.7 $\pm$ 2.5\\
ComQA & 20.5 & 20.5 & 20.5 & 53.3 & 61.6 & 63.2 & 33.0 $\pm$ 2.4 & 36.0 $\pm$ 1.8 & 39.1 $\pm$ 1.2\\
\bottomrule
 \end{tabular}
 \end{adjustbox}
  \caption{Fine-tuning results for question answering tasks.}
  \label{tbla2}
\end{table*}

\clearpage
\newpage
\begin{table*}[t]
\centering
\begin{adjustbox}{max width=\textwidth}
  \begin{tabular}{ l  l l l | l l l | l l l}
  \toprule
\multirow{3}{*}{\textbf{Task}} %& \multicolumn{9}{c }{\textbf{Data regime}}\\ \cline{2-10}
& \multicolumn{6}{c}{\textsc{Full}} & \multicolumn{3}{c}{\textsc{Limited}} \\ \cmidrule(lr){2-7} \cmidrule(lr){8-10}  
& \multicolumn{3}{c}{\emph{frozen BERT}} &  \multicolumn{3}{c}{\emph{unfrozen BERT}} & \multicolumn{3}{c}{\emph{unfrozen BERT}} \\ 
\cmidrule(lr){2-4} \cmidrule(lr){5-7}  
& \multicolumn{1}{c}{t = 2} & \multicolumn{1}{c}{t = 4} & \multicolumn{1}{c}{t = 6} & \multicolumn{1}{c}{t = 2} & \multicolumn{1}{c}{t = 4} & \multicolumn{1}{c}{t = 6} & \multicolumn{1}{c}{t = 2} & \multicolumn{1}{c}{t = 4} & \multicolumn{1}{c}{t = 6}  \\
\midrule
CCG & 39.7 & 44.9 & 48.1 & 95.2 & 95.5 & 95.6 & 11.1 $\pm$ 6.1 & 45.2 $\pm$ 3.9 & 53.2 $\pm$ 1.6\\
POS-PTB & 61.7 & 74.0 & 76.4 & 96.6 & 96.6 & 96.7 & 46.5 $\pm$ 2.8 & 80.5 $\pm$ 1.1 & 85.1 $\pm$ 0.9\\
POS-EWT & 33.5 & 46.0 & 49.1 & 96.2 & 96.5 & 96.6 & 65.4 $\pm$ 3.0 & 86.8 $\pm$ 0.6 & 89.3 $\pm$ 0.4\\
Parent & 37.9 & 58.1 & 61.5 & 95.1 & 95.3 & 95.4 & 61.1 $\pm$ 4.0 & 77.0 $\pm$ 1.0 & 81.9 $\pm$ 0.9\\
GParent & 35.0 & 41.9 & 43.4 & 91.1 & 91.7 & 91.9 & 41.1 $\pm$ 1.4 & 58.0 $\pm$ 1.7 & 62.8 $\pm$ 1.3\\
GGParent & 25.9 & 30.9 & 31.7 & 88.3 & 89.3 & 89.5 & 25.6 $\pm$ 3.1 & 37.9 $\pm$ 1.7 & 43.3 $\pm$ 1.7\\
ST & 51.2 & 66.1 & 69.2 & 95.5 & 95.7 & 95.8 & 38.6 $\pm$ 1.1 & 71.3 $\pm$ 1.6 & 76.7 $\pm$ 0.9\\
Chunk & 11.9 & 16.6 & 18.4 & 96.4 & 96.8 & 97.1 & 68.1 $\pm$ 2.4 & 85.0 $\pm$ 0.7 & 87.7 $\pm$ 0.5\\
NER & 4.7 & 7.7 & 9.2 & 93.8 & 94.3 & 94.7 & 58.4 $\pm$ 7.3 & 73.5 $\pm$ 1.6 & 77.4 $\pm$ 1.5\\
GED & 16.8 & 18.4 & 18.8 & 44.2 & 46.9 & 46.6 & 17.3 $\pm$ 1.2 & 27.4 $\pm$ 1.4 & 29.1 $\pm$ 1.3\\
Conj & 14.7 & 19.8 & 21.1 & 88.6 & 89.9 & 89.4 & 40.6 $\pm$ 6.0 & 69.2 $\pm$ 2.4 & 73.3 $\pm$ 1.6\\
\bottomrule
 \end{tabular}
 \end{adjustbox}
  \caption{Fine-tuning results for sequence labeling tasks.}
  \label{tbla3}
\end{table*}

\clearpage
\newpage
\global\pdfpageattr\expandafter{\the\pdfpageattr/Rotate 90}
\begin{landscape}
\begin{table}[t]
\centering
\begin{adjustbox}{max width=2\textwidth}
\begin{tabular}{ l  l l l l  l l l l  l l l }
\toprule
\textbf{Task} & CoLA & SST-2 & MRPC & STS-B & QQP & MNLI & QNLI & RTE & WNLI & SNLI & SciTail \\
\midrule
CoLA & \textcolor{orange}{51.0} & \textcolor{blue}{92.2} & \textcolor{blue}{86.6} & \textcolor{blue}{86.4} & \textbf{\textcolor{blue}{87.5}} & 84.2 & 91.4 & 60.3 & \textbf{\textcolor{blue}{54.9}} & 90.5 & 93.8\\
SST-2 & \textcolor{blue}{54.2} & \textcolor{orange}{91.9} & \textcolor{blue}{84.2} & \textcolor{blue}{86.9} & 87.0 & 84.1 & 91.3 & 56.0 & \textcolor{blue}{53.5} & \textcolor{blue}{90.9} & 93.5\\
MRPC & 51.0 & \textcolor{blue}{92.3} & \textcolor{orange}{84.0} & \textcolor{blue}{87.1} & 87.1 & \textcolor{blue}{84.4} & 91.3 & \textcolor{blue}{61.7} & \textcolor{blue}{47.9} & \textcolor{blue}{90.9} & 93.5\\
STS-B & 48.8 & 91.9 & \textcolor{blue}{87.3} & \textcolor{orange}{85.9} & 86.4 & 84.0 & 90.4 & \textcolor{blue}{65.0} & 35.2 & \textcolor{blue}{90.9} & 92.1\\
QQP & 49.4 & \textcolor{blue}{92.0} & \textbf{\textcolor{blue}{87.7}} & \textcolor{blue}{88.5} & \textcolor{orange}{87.3} & 84.2 & 90.7 & \textcolor{blue}{61.7} & 36.6 & \textcolor{blue}{90.9} & 92.9\\
MNLI & 50.0 & \textbf{\textcolor{blue}{93.5}} & \textcolor{blue}{87.6} & \textcolor{blue}{87.0} & 87.1 & \textcolor{orange}{84.2} & \textbf{\textcolor{blue}{91.5}} & \textbf{\textcolor{blue}{77.6}} & 40.8 & \textbf{\textcolor{blue}{91.2}} & \textbf{\textcolor{blue}{95.6}}\\
QNLI & 49.9 & \textcolor{blue}{92.5} & \textcolor{blue}{86.6} & \textbf{\textcolor{blue}{88.6}} & 86.6 & \textcolor{blue}{84.4} & \textcolor{orange}{91.4} & \textcolor{blue}{70.4} & 38.0 & \textcolor{blue}{91.1} & \textcolor{blue}{94.5}\\
\bottomrule
\end{tabular}
\end{adjustbox}
\caption{In-class transfer results for classification/regression tasks in the \smallsc{Full $\rightarrow$ Full} regime.}
\label{tbla4a}
\end{table}
\end{landscape}
\clearpage
\newpage
\global\pdfpageattr\expandafter{\the\pdfpageattr/Rotate 90}
\begin{landscape}
\begin{table}[t]
\centering
\begin{adjustbox}{max width=2\textwidth}
\begin{tabular}{ l l l l l  l l l l  l l l }
\toprule
\textbf{Task} & CoLA & SST-2 & MRPC & STS-B & QQP & MNLI & QNLI & RTE & WNLI & SNLI & SciTail \\
\midrule
CoLA & \textcolor{orange}{4.7 $\pm$ 8.2} & 74.4 $\pm$ 5.9 & \textcolor{blue}{82.0 $\pm$ 0.5} & \textcolor{blue}{32.7 $\pm$ 10.6} & \textcolor{blue}{38.2 $\pm$ 28.8} & \textcolor{blue}{39.3 $\pm$ 2.6} & \textcolor{blue}{66.7 $\pm$ 6.1} & \textcolor{blue}{56.4 $\pm$ 2.7} & 40.1 $\pm$ 8.3 & \textcolor{blue}{47.4 $\pm$ 2.8} & \textcolor{blue}{68.6 $\pm$ 15.7}\\
SST-2 & 1.3 $\pm$ 2.8 & \textcolor{orange}{77.5 $\pm$ 6.3} & 81.9 $\pm$ 0.7 & 29.1 $\pm$ 12.7 & \textcolor{blue}{33.1 $\pm$ 23.2} & \textcolor{blue}{43.6 $\pm$ 3.4} & \textcolor{blue}{66.4 $\pm$ 7.0} & \textcolor{blue}{55.0 $\pm$ 2.8} & 39.7 $\pm$ 5.6 & \textcolor{blue}{49.3 $\pm$ 2.8} & \textcolor{blue}{64.5 $\pm$ 14.9}\\
MRPC & 1.2 $\pm$ 4.3 & 68.4 $\pm$ 11.3 & \textcolor{orange}{81.9 $\pm$ 0.7} & \textbf{\textcolor{blue}{71.2 $\pm$ 6.7}} & \textcolor{blue}{54.2 $\pm$ 22.0} & \textcolor{blue}{46.3 $\pm$ 2.0} & \textcolor{blue}{73.5 $\pm$ 1.6} & \textcolor{blue}{59.2 $\pm$ 1.7} & 38.7 $\pm$ 6.4 & \textcolor{blue}{51.9 $\pm$ 2.5} & \textcolor{blue}{84.7 $\pm$ 1.0}\\
STS-B & 2.3 $\pm$ 5.2 & 75.8 $\pm$ 7.4 & \textbf{\textcolor{blue}{84.6 $\pm$ 0.5}} & \textcolor{orange}{29.9 $\pm$ 10.5} & \textbf{\textcolor{blue}{67.5 $\pm$ 1.4}} & \textcolor{blue}{49.2 $\pm$ 1.2} & \textcolor{blue}{76.7 $\pm$ 0.5} & \textcolor{blue}{62.2 $\pm$ 1.9} & \textbf{\textcolor{blue}{44.6 $\pm$ 8.5}} & \textcolor{blue}{55.4 $\pm$ 1.7} & \textcolor{blue}{86.4 $\pm$ 1.1}\\
QQP & \textbf{\textcolor{blue}{7.7 $\pm$ 9.0}} & \textcolor{blue}{82.0 $\pm$ 2.3} & \textcolor{blue}{83.5 $\pm$ 1.2} & \textcolor{blue}{67.4 $\pm$ 8.3} & \textcolor{orange}{25.7 $\pm$ 25.1} & \textcolor{blue}{52.4 $\pm$ 3.0} & \textbf{\textcolor{blue}{77.1 $\pm$ 1.3}} & \textcolor{blue}{62.8 $\pm$ 2.2} & 36.4 $\pm$ 6.5 & \textcolor{blue}{56.4 $\pm$ 2.8} & \textbf{\textcolor{blue}{88.2 $\pm$ 1.4}}\\
\bottomrule
\end{tabular}
\end{adjustbox}
\caption{In-class transfer results for classification/regression tasks in the \smallsc{Full $\rightarrow$ Limited} regime.}
\label{tbla4b}
\end{table}
\end{landscape}
\clearpage
\newpage
\global\pdfpageattr\expandafter{\the\pdfpageattr/Rotate 90}
\begin{landscape}
\begin{table}[t]
\centering
\begin{adjustbox}{max width=2\textwidth}
\begin{tabular}{ l l l l l  l l l l  l l l }
\toprule
\textbf{Task} & CoLA & SST-2 & MRPC & STS-B & QQP & MNLI & QNLI & RTE & WNLI & SNLI & SciTail \\
\midrule
CoLA & \textcolor{orange}{4.7 $\pm$ 8.2} & 74.8 $\pm$ 6.1 & 81.9 $\pm$ 0.7 & 24.1 $\pm$ 10.3 & \textcolor{blue}{28.0 $\pm$ 27.3} & 38.4 $\pm$ 3.2 & 62.3 $\pm$ 9.5 & \textcolor{blue}{54.8 $\pm$ 3.0} & 43.7 $\pm$ 6.4 & \textcolor{blue}{47.1 $\pm$ 3.9} & \textcolor{blue}{65.2 $\pm$ 13.6}\\
SST-2 & 4.2 $\pm$ 7.3 & \textcolor{orange}{77.5 $\pm$ 6.3} & 81.9 $\pm$ 0.7 & 27.9 $\pm$ 10.8 & \textcolor{blue}{33.4 $\pm$ 26.5} & \textcolor{blue}{39.1 $\pm$ 3.4} & \textcolor{blue}{63.8 $\pm$ 8.9} & \textcolor{blue}{55.9 $\pm$ 3.5} & 43.9 $\pm$ 6.4 & \textcolor{blue}{47.8 $\pm$ 3.6} & \textcolor{blue}{65.3 $\pm$ 13.9}\\
MRPC & 2.5 $\pm$ 5.2 & 75.2 $\pm$ 8.1 & \textcolor{orange}{81.9 $\pm$ 0.7} & \textcolor{blue}{45.2 $\pm$ 11.8} & \textcolor{blue}{40.0 $\pm$ 28.3} & \textcolor{blue}{41.2 $\pm$ 3.8} & \textcolor{blue}{68.8 $\pm$ 5.8} & \textcolor{blue}{57.2 $\pm$ 3.9} & 41.7 $\pm$ 7.9 & \textcolor{blue}{51.3 $\pm$ 2.7} & \textcolor{blue}{73.1 $\pm$ 14.8}\\
STS-B & \textbf{\textcolor{blue}{6.7 $\pm$ 8.1}} & 76.7 $\pm$ 6.8 & \textcolor{blue}{82.0 $\pm$ 0.7} & \textcolor{orange}{29.9 $\pm$ 10.5} & \textbf{\textcolor{blue}{43.8 $\pm$ 23.2}} & \textbf{\textcolor{blue}{43.9 $\pm$ 2.2}} & \textbf{\textcolor{blue}{73.2 $\pm$ 1.1}} & \textcolor{blue}{58.6 $\pm$ 2.6} & 39.2 $\pm$ 6.1 & \textbf{\textcolor{blue}{51.8 $\pm$ 2.7}} & \textbf{\textcolor{blue}{79.3 $\pm$ 6.6}}\\
QQP & 3.2 $\pm$ 5.4 & 76.6 $\pm$ 8.3 & \textcolor{blue}{82.1 $\pm$ 0.8} & \textcolor{blue}{35.7 $\pm$ 12.1} & \textcolor{orange}{25.7 $\pm$ 25.1} & \textcolor{blue}{40.4 $\pm$ 4.1} & \textcolor{blue}{65.5 $\pm$ 8.1} & \textcolor{blue}{55.5 $\pm$ 3.9} & 39.7 $\pm$ 8.6 & \textcolor{blue}{49.8 $\pm$ 2.7} & \textcolor{blue}{69.3 $\pm$ 16.2}\\
\bottomrule
\end{tabular}
\end{adjustbox}
\caption{In-class transfer results for classification/regression tasks in the \smallsc{Limited $\rightarrow$ Limited} regime.}
\label{tbla4c}
\end{table}
\end{landscape}
\clearpage
\newpage
\global\pdfpageattr\expandafter{\the\pdfpageattr/Rotate 90}
\begin{landscape}
\begin{table}[t]
\centering
\begin{adjustbox}{max width=2\textwidth}
\begin{tabular}{ l l l l l  l l l l  l l l }
\toprule
\textbf{Task} & SQuAD-1 & SQuAD-2 & NewsQA & HotpotQA & BoolQ & DROP & WikiHop & DuoRC-p & DuoRC-s & CQ & ComQA \\
\midrule
SQuAD-1 & \textcolor{orange}{87.9} & \textcolor{blue}{73.4} & \textcolor{blue}{65.5} & \textcolor{blue}{70.1} & \textcolor{blue}{71.0} & \textcolor{blue}{26.9} & \textcolor{blue}{63.7} & \textcolor{blue}{51.1} & 62.9 & \textcolor{blue}{45.2} & \textcolor{blue}{64.8}\\
SQuAD-2 & 87.8 & \textcolor{orange}{71.9} & \textbf{\textcolor{blue}{66.3}} & \textbf{\textcolor{blue}{70.6}} & \textbf{\textcolor{blue}{74.3}} & \textbf{\textcolor{blue}{27.7}} & \textcolor{blue}{63.6} & \textcolor{blue}{51.2} & 62.9 & \textbf{\textcolor{blue}{45.4}} & \textcolor{blue}{64.4}\\
NewsQA & \textbf{\textcolor{blue}{89.0}} & \textbf{\textcolor{blue}{73.8}} & \textcolor{orange}{64.1} & \textcolor{blue}{69.7} & \textcolor{blue}{73.0} & \textcolor{blue}{27.4} & \textcolor{blue}{63.6} & \textcolor{blue}{50.7} & 61.8 & \textcolor{blue}{41.2} & \textbf{\textcolor{blue}{65.3}}\\
HotpotQA & \textcolor{blue}{88.6} & \textcolor{blue}{72.8} & \textcolor{blue}{64.8} & \textcolor{orange}{67.9} & \textcolor{blue}{73.1} & \textcolor{blue}{26.1} & \textbf{\textcolor{blue}{64.2}} & 50.2 & 62.0 & \textcolor{blue}{45.3} & \textcolor{blue}{63.3}\\
BoolQ & 87.8 & 70.3 & \textcolor{blue}{64.5} & \textcolor{blue}{68.0} & \textcolor{orange}{65.7} & 22.2 & \textcolor{blue}{63.0} & \textcolor{blue}{50.8} & 62.1 & \textcolor{blue}{33.0} & \textcolor{blue}{63.6}\\
DROP & \textcolor{blue}{88.1} & 71.8 & \textcolor{blue}{65.6} & \textcolor{blue}{69.6} & \textcolor{blue}{69.0} & \textcolor{orange}{22.4} & \textcolor{blue}{63.7} & \textcolor{blue}{50.8} & 63.0 & \textcolor{blue}{41.5} & \textcolor{blue}{65.2}\\
WikiHop & 87.4 & 69.2 & 63.7 & \textcolor{blue}{68.4} & \textcolor{blue}{68.3} & 21.8 & \textcolor{orange}{62.8} & 50.1 & 61.2 & \textcolor{blue}{43.5} & \textbf{\textcolor{blue}{65.3}}\\
\bottomrule
\end{tabular}
\end{adjustbox}
\caption{In-class transfer results for question answering tasks in the \smallsc{Full $\rightarrow$ Full} regime.}
\label{tbla5a}
\end{table}
\end{landscape}
\clearpage
\newpage
\global\pdfpageattr\expandafter{\the\pdfpageattr/Rotate 90}
\begin{landscape}
\begin{table}[t]
\centering
\begin{adjustbox}{max width=2\textwidth}
\begin{tabular}{ l l l l l  l l l l  l l l }
\toprule
\textbf{Task} & SQuAD-1 & SQuAD-2 & NewsQA & HotpotQA & BoolQ & DROP & WikiHop & DuoRC-p & DuoRC-s & CQ & ComQA \\
\midrule
SQuAD-1 & \textcolor{orange}{26.8 $\pm$ 6.0} & \textbf{\textcolor{blue}{57.4 $\pm$ 1.1}} & \textcolor{blue}{57.1 $\pm$ 0.4} & \textcolor{blue}{50.9 $\pm$ 0.5} & 62.2 $\pm$ 0.0 & 16.8 $\pm$ 0.8 & \textcolor{blue}{38.1 $\pm$ 1.1} & \textcolor{blue}{50.2 $\pm$ 0.9} & \textcolor{blue}{59.8 $\pm$ 1.0} & \textcolor{blue}{45.0 $\pm$ 2.1} & \textcolor{blue}{46.6 $\pm$ 1.0}\\
SQuAD-2 & \textbf{\textcolor{blue}{86.5 $\pm$ 0.3}} & \textcolor{orange}{50.1 $\pm$ 0.1} & \textbf{\textcolor{blue}{57.2 $\pm$ 0.4}} & \textbf{\textcolor{blue}{51.2 $\pm$ 0.5}} & 62.2 $\pm$ 0.0 & \textbf{\textcolor{blue}{26.0 $\pm$ 4.0}} & \textcolor{blue}{37.7 $\pm$ 1.0} & \textbf{\textcolor{blue}{51.0 $\pm$ 1.1}} & \textbf{\textcolor{blue}{60.6 $\pm$ 0.8}} & \textcolor{blue}{44.5 $\pm$ 1.7} & \textcolor{blue}{46.3 $\pm$ 0.7}\\
NewsQA & \textcolor{blue}{79.4 $\pm$ 0.7} & \textcolor{blue}{55.8 $\pm$ 0.9} & \textcolor{orange}{28.8 $\pm$ 4.9} & \textcolor{blue}{48.0 $\pm$ 0.5} & 62.2 $\pm$ 0.0 & 16.0 $\pm$ 1.8 & \textcolor{blue}{38.1 $\pm$ 0.6} & \textcolor{blue}{49.9 $\pm$ 0.6} & \textcolor{blue}{57.9 $\pm$ 0.7} & \textcolor{blue}{43.0 $\pm$ 2.3} & \textcolor{blue}{47.4 $\pm$ 1.1}\\
HotpotQA & \textcolor{blue}{78.4 $\pm$ 0.4} & \textcolor{blue}{54.1 $\pm$ 0.9} & \textcolor{blue}{52.8 $\pm$ 0.4} & \textcolor{orange}{23.3 $\pm$ 4.0} & 62.2 $\pm$ 0.1 & \textcolor{blue}{20.0 $\pm$ 2.8} & \textbf{\textcolor{blue}{39.4 $\pm$ 0.8}} & \textcolor{blue}{48.7 $\pm$ 0.8} & \textcolor{blue}{55.5 $\pm$ 1.2} & \textbf{\textcolor{blue}{46.9 $\pm$ 2.0}} & \textbf{\textcolor{blue}{47.9 $\pm$ 1.0}}\\
BoolQ & 26.6 $\pm$ 6.8 & 50.1 $\pm$ 0.0 & 26.3 $\pm$ 3.9 & \textcolor{blue}{31.0 $\pm$ 4.1} & \textcolor{orange}{62.2 $\pm$ 0.0} & 15.3 $\pm$ 12.2 & 18.9 $\pm$ 3.3 & 41.2 $\pm$ 1.2 & 34.5 $\pm$ 3.3 & \textcolor{blue}{31.9 $\pm$ 2.0} & 38.9 $\pm$ 1.4\\
\bottomrule
\end{tabular}
\end{adjustbox}
\caption{In-class transfer results for question answering tasks in the \smallsc{Full $\rightarrow$ Limited} regime.}
\label{tbla5b}
\end{table}
\end{landscape}
\clearpage
\newpage
\global\pdfpageattr\expandafter{\the\pdfpageattr/Rotate 90}
\begin{landscape}
\begin{table}[t]
\centering
\begin{adjustbox}{max width=2\textwidth}
\begin{tabular}{ l l l l l  l l l l  l l l }
\toprule
\textbf{Task} & SQuAD-1 & SQuAD-2 & NewsQA & HotpotQA & BoolQ & DROP & WikiHop & DuoRC-p & DuoRC-s & CQ & ComQA \\
\midrule
SQuAD-1 & \textcolor{orange}{26.8 $\pm$ 6.0} & 42.8 $\pm$ 2.9 & \textcolor{blue}{35.3 $\pm$ 2.5} & \textcolor{blue}{31.5 $\pm$ 2.1} & 62.2 $\pm$ 0.0 & 9.5 $\pm$ 0.5 & \textcolor{blue}{27.9 $\pm$ 3.2} & \textcolor{blue}{44.6 $\pm$ 0.7} & \textcolor{blue}{42.9 $\pm$ 1.8} & \textcolor{blue}{33.2 $\pm$ 1.8} & \textcolor{blue}{39.7 $\pm$ 1.0}\\
SQuAD-2 & \textcolor{blue}{48.7 $\pm$ 1.6} & \textcolor{orange}{50.1 $\pm$ 0.1} & \textcolor{blue}{39.9 $\pm$ 1.0} & \textcolor{blue}{34.3 $\pm$ 3.2} & 62.2 $\pm$ 0.0 & 17.8 $\pm$ 5.6 & \textcolor{blue}{29.5 $\pm$ 2.2} & \textcolor{blue}{45.0 $\pm$ 0.7} & \textcolor{blue}{46.6 $\pm$ 1.7} & \textcolor{blue}{32.2 $\pm$ 2.4} & \textcolor{blue}{39.5 $\pm$ 1.0}\\
NewsQA & \textbf{\textcolor{blue}{63.8 $\pm$ 1.1}} & 45.8 $\pm$ 1.7 & \textcolor{orange}{28.8 $\pm$ 4.9} & \textbf{\textcolor{blue}{42.3 $\pm$ 0.5}} & 62.2 $\pm$ 0.0 & 17.2 $\pm$ 3.8 & \textcolor{blue}{33.5 $\pm$ 0.8} & \textbf{\textcolor{blue}{47.0 $\pm$ 0.7}} & \textbf{\textcolor{blue}{51.0 $\pm$ 1.0}} & \textcolor{blue}{38.0 $\pm$ 2.3} & \textbf{\textcolor{blue}{42.7 $\pm$ 1.1}}\\
HotpotQA & \textcolor{blue}{59.4 $\pm$ 1.0} & 46.5 $\pm$ 1.4 & \textbf{\textcolor{blue}{43.6 $\pm$ 0.8}} & \textcolor{orange}{23.3 $\pm$ 4.0} & 62.2 $\pm$ 0.0 & 17.5 $\pm$ 7.8 & \textbf{\textcolor{blue}{35.0 $\pm$ 0.8}} & \textcolor{blue}{46.6 $\pm$ 0.6} & \textcolor{blue}{50.2 $\pm$ 0.9} & \textbf{\textcolor{blue}{39.7 $\pm$ 1.7}} & \textbf{\textcolor{blue}{42.7 $\pm$ 1.4}}\\
BoolQ & \textcolor{blue}{32.4 $\pm$ 7.8} & 50.0 $\pm$ 0.1 & 25.3 $\pm$ 2.8 & \textcolor{blue}{26.0 $\pm$ 4.3} & \textcolor{orange}{62.2 $\pm$ 0.0} & \textbf{\textcolor{blue}{49.1 $\pm$ 14.4}} & 23.1 $\pm$ 4.4 & \textcolor{blue}{42.1 $\pm$ 0.7} & 35.4 $\pm$ 5.6 & \textcolor{blue}{31.4 $\pm$ 2.5} & 38.7 $\pm$ 1.1\\
\bottomrule
\end{tabular}
\end{adjustbox}
\caption{In-class transfer results for question answering tasks in the \smallsc{Limited $\rightarrow$ Limited} regime.}
\label{tbla5c}
\end{table}
\end{landscape}
\clearpage
\newpage
\global\pdfpageattr\expandafter{\the\pdfpageattr/Rotate 90}
\begin{landscape}
\begin{table}[t]
\centering
\begin{adjustbox}{max width=2\textwidth}
\begin{tabular}{ l l l l l  l l l l  l l l }
\toprule
\textbf{Task} & CCG & POS-PTB & POS-EWT & Parent & GParent & GGParent & ST & Chunk & NER & GED & Conj \\
\midrule
CCG & \textcolor{orange}{95.6} & 96.7 & 96.4 & 95.3 & 91.8 & \textcolor{blue}{89.6} & 95.8 & \textcolor{blue}{97.7} & 94.0 & 45.8 & \textcolor{blue}{90.3}\\
POS-PTB & \textbf{\textcolor{blue}{95.7}} & \textcolor{orange}{96.7} & \textbf{\textcolor{blue}{96.7}} & 95.3 & 91.7 & 89.1 & 95.7 & 97.0 & 94.6 & 46.5 & \textcolor{blue}{90.2}\\
POS-EWT & 95.6 & 96.7 & \textcolor{orange}{96.6} & \textbf{\textcolor{blue}{95.5}} & 91.9 & 89.3 & 95.8 & 97.0 & 94.6 & 46.1 & \textcolor{blue}{89.9}\\
Parent & 95.6 & 96.7 & 96.6 & \textcolor{orange}{95.4} & 91.9 & \textcolor{blue}{89.8} & 95.8 & \textbf{\textcolor{blue}{98.0}} & 94.5 & 46.6 & \textcolor{blue}{90.3}\\
GParent & 95.6 & 96.7 & 96.6 & 95.1 & \textcolor{orange}{91.9} & \textbf{\textcolor{blue}{90.0}} & 95.8 & \textcolor{blue}{97.6} & 94.6 & 46.5 & \textbf{\textcolor{blue}{91.0}}\\
GGParent & 95.5 & 96.6 & 96.5 & 95.4 & 91.9 & \textcolor{orange}{89.5} & 95.8 & \textcolor{blue}{97.5} & 94.5 & 46.5 & \textcolor{blue}{90.8}\\
ST & 95.5 & 96.6 & 96.5 & 95.1 & 91.6 & 89.3 & \textcolor{orange}{95.8} & 96.9 & \textbf{\textcolor{blue}{94.9}} & 46.2 & 88.7\\
Chunk & 95.6 & 96.7 & 96.5 & 95.2 & 91.8 & 89.5 & 95.7 & \textcolor{orange}{97.1} & 94.6 & 46.4 & \textcolor{blue}{89.7}\\
NER & 95.4 & 96.7 & 96.6 & 95.2 & 91.7 & 89.1 & 95.8 & 97.0 & \textcolor{orange}{94.7} & \textbf{\textcolor{blue}{47.3}} & \textcolor{blue}{90.3}\\
GED & 95.5 & 96.7 & 96.6 & 95.2 & 91.7 & 89.3 & 95.8 & 97.0 & 94.7 & \textcolor{orange}{46.6} & \textcolor{blue}{90.2}\\
Conj & 95.4 & 96.7 & 96.6 & 95.4 & 91.9 & \textcolor{blue}{89.7} & 95.8 & 97.0 & 94.5 & 46.2 & \textcolor{orange}{89.4}\\
\bottomrule
\end{tabular}
\end{adjustbox}
\caption{In-class transfer results for sequence labeling tasks in the \smallsc{Full $\rightarrow$ Full} regime.}
\label{tbla6a}
\end{table}
\end{landscape}
\clearpage
\newpage
\global\pdfpageattr\expandafter{\the\pdfpageattr/Rotate 90}
\begin{landscape}
\begin{table}[t]
\centering
\begin{adjustbox}{max width=2\textwidth}
\begin{tabular}{ l l l l l  l l l l  l l l }
\toprule
\textbf{Task} & CCG & POS-PTB & POS-EWT & Parent & GParent & GGParent & ST & Chunk & NER & GED & Conj \\
\midrule
CCG & \textcolor{orange}{53.2 $\pm$ 1.6} & \textcolor{blue}{89.8 $\pm$ 0.8} & \textcolor{blue}{91.9 $\pm$ 0.2} & \textcolor{blue}{87.1 $\pm$ 1.3} & \textcolor{blue}{74.5 $\pm$ 0.5} & \textcolor{blue}{54.0 $\pm$ 1.3} & \textcolor{blue}{84.0 $\pm$ 0.8} & \textcolor{blue}{92.9 $\pm$ 0.1} & 67.9 $\pm$ 3.3 & 24.3 $\pm$ 1.4 & 73.2 $\pm$ 1.3\\
POS-PTB & \textcolor{blue}{72.0 $\pm$ 0.5} & \textcolor{orange}{85.1 $\pm$ 0.9} & \textbf{\textcolor{blue}{93.9 $\pm$ 0.2}} & \textcolor{blue}{87.7 $\pm$ 0.7} & \textcolor{blue}{68.4 $\pm$ 1.0} & \textcolor{blue}{49.5 $\pm$ 1.1} & \textbf{\textcolor{blue}{86.3 $\pm$ 0.2}} & \textcolor{blue}{91.2 $\pm$ 0.5} & \textcolor{blue}{83.3 $\pm$ 1.2} & 28.8 $\pm$ 0.9 & 69.5 $\pm$ 2.0\\
POS-EWT & \textcolor{blue}{68.2 $\pm$ 0.7} & \textcolor{blue}{88.5 $\pm$ 0.6} & \textcolor{orange}{89.3 $\pm$ 0.4} & \textcolor{blue}{86.4 $\pm$ 1.0} & \textcolor{blue}{66.2 $\pm$ 1.0} & \textcolor{blue}{47.5 $\pm$ 1.2} & \textcolor{blue}{83.4 $\pm$ 0.8} & \textcolor{blue}{91.8 $\pm$ 0.3} & \textcolor{blue}{81.3 $\pm$ 1.3} & 29.1 $\pm$ 0.9 & 70.9 $\pm$ 2.7\\
Parent & \textcolor{blue}{66.2 $\pm$ 1.0} & \textcolor{blue}{88.5 $\pm$ 0.9} & \textcolor{blue}{92.6 $\pm$ 0.3} & \textcolor{orange}{81.9 $\pm$ 0.9} & \textcolor{blue}{75.8 $\pm$ 0.7} & \textcolor{blue}{55.4 $\pm$ 1.7} & \textcolor{blue}{82.4 $\pm$ 0.7} & \textbf{\textcolor{blue}{94.3 $\pm$ 0.3}} & \textcolor{blue}{78.3 $\pm$ 4.0} & 28.7 $\pm$ 0.9 & \textcolor{blue}{76.5 $\pm$ 3.8}\\
GParent & \textcolor{blue}{64.5 $\pm$ 3.0} & \textcolor{blue}{87.2 $\pm$ 1.0} & \textcolor{blue}{90.8 $\pm$ 0.3} & \textbf{\textcolor{blue}{90.5 $\pm$ 0.2}} & \textcolor{orange}{62.8 $\pm$ 1.3} & \textbf{\textcolor{blue}{77.4 $\pm$ 0.6}} & \textcolor{blue}{81.6 $\pm$ 0.6} & \textcolor{blue}{92.0 $\pm$ 0.3} & 76.4 $\pm$ 1.9 & 24.2 $\pm$ 1.7 & \textcolor{blue}{83.4 $\pm$ 0.6}\\
\bottomrule
\end{tabular}
\end{adjustbox}
\caption{In-class transfer results for sequence labeling tasks in the \smallsc{Full $\rightarrow$ Limited} regime.}
\label{tbla6b}
\end{table}
\end{landscape}
\clearpage
\newpage
\global\pdfpageattr\expandafter{\the\pdfpageattr/Rotate 90}
\begin{landscape}
\begin{table}[t]
\centering
\begin{adjustbox}{max width=2\textwidth}
\begin{tabular}{ l l l l l  l l l l  l l l }
\toprule
\textbf{Task} & CCG & POS-PTB & POS-EWT & Parent & GParent & GGParent & ST & Chunk & NER & GED & Conj \\
\midrule
CCG & \textcolor{orange}{53.2 $\pm$ 1.6} & \textcolor{blue}{88.6 $\pm$ 0.4} & \textcolor{blue}{90.8 $\pm$ 0.4} & \textcolor{blue}{85.7 $\pm$ 0.3} & \textcolor{blue}{64.9 $\pm$ 1.4} & \textcolor{blue}{44.2 $\pm$ 3.8} & \textcolor{blue}{80.0 $\pm$ 1.4} & \textcolor{blue}{89.3 $\pm$ 0.2} & 63.3 $\pm$ 4.7 & 27.8 $\pm$ 2.1 & 66.3 $\pm$ 4.5\\
POS-PTB & \textbf{\textcolor{blue}{68.2 $\pm$ 0.8}} & \textcolor{orange}{85.1 $\pm$ 0.9} & \textbf{\textcolor{blue}{92.0 $\pm$ 0.1}} & \textcolor{blue}{85.7 $\pm$ 0.5} & \textcolor{blue}{65.3 $\pm$ 1.7} & \textcolor{blue}{43.6 $\pm$ 2.4} & \textbf{\textcolor{blue}{83.2 $\pm$ 0.6}} & \textcolor{blue}{89.5 $\pm$ 0.3} & 76.9 $\pm$ 2.2 & 28.5 $\pm$ 1.8 & 61.5 $\pm$ 9.9\\
POS-EWT & \textcolor{blue}{66.7 $\pm$ 0.7} & \textcolor{blue}{88.7 $\pm$ 1.3} & \textcolor{orange}{89.3 $\pm$ 0.4} & \textcolor{blue}{86.0 $\pm$ 0.9} & \textcolor{blue}{65.0 $\pm$ 1.4} & 43.0 $\pm$ 3.3 & \textcolor{blue}{82.6 $\pm$ 0.9} & \textcolor{blue}{90.2 $\pm$ 0.5} & \textcolor{blue}{77.7 $\pm$ 2.5} & 27.1 $\pm$ 0.8 & 57.7 $\pm$ 7.6\\
Parent & \textcolor{blue}{66.0 $\pm$ 2.0} & \textcolor{blue}{88.5 $\pm$ 0.9} & \textcolor{blue}{91.5 $\pm$ 0.2} & \textcolor{orange}{81.9 $\pm$ 0.9} & \textcolor{blue}{68.5 $\pm$ 0.8} & \textcolor{blue}{47.6 $\pm$ 2.5} & \textcolor{blue}{80.5 $\pm$ 1.2} & \textbf{\textcolor{blue}{90.3 $\pm$ 0.3}} & 74.0 $\pm$ 2.6 & 29.1 $\pm$ 2.3 & 66.7 $\pm$ 4.0\\
GParent & \textcolor{blue}{63.7 $\pm$ 1.4} & \textcolor{blue}{87.9 $\pm$ 0.6} & \textcolor{blue}{90.7 $\pm$ 0.4} & \textcolor{blue}{86.4 $\pm$ 0.5} & \textcolor{orange}{62.8 $\pm$ 1.3} & \textbf{\textcolor{blue}{58.1 $\pm$ 1.5}} & \textcolor{blue}{80.0 $\pm$ 0.9} & \textcolor{blue}{89.9 $\pm$ 0.3} & 70.1 $\pm$ 3.6 & \textcolor{blue}{29.3 $\pm$ 1.6} & \textcolor{blue}{75.9 $\pm$ 2.7}\\
\bottomrule
\end{tabular}
\end{adjustbox}
\caption{In-class transfer results for sequence labeling tasks in the \smallsc{Limited $\rightarrow$ Limited} regime.}
\label{tbla6c}
\end{table}
\end{landscape}
\clearpage
\newpage
\global\pdfpageattr\expandafter{\the\pdfpageattr/Rotate 90}
\begin{landscape}
\begin{table}[t]
\centering
\begin{adjustbox}{max width=2\textwidth}
\begin{tabular}{ l l l l l  l l l l  l l l }
\toprule
\textbf{Task} & CoLA & SST-2 & MRPC & STS-B & QQP & MNLI & QNLI & RTE & WNLI & SNLI & SciTail \\
\midrule
\emph{Baseline} & \textcolor{orange}{51.0} & \textcolor{orange}{91.9} & \textcolor{orange}{84.0} & \textcolor{orange}{85.9} & \textcolor{orange}{87.3} & \textcolor{orange}{84.2} & \textcolor{orange}{91.4} & \textcolor{orange}{60.6} & \textcolor{orange}{45.1} & \textcolor{orange}{90.7} & \textcolor{orange}{93.9}\\
SQuAD-1 & \textcolor{blue}{52.4} & \textcolor{blue}{92.1} & \textcolor{blue}{87.0} & \textbf{\textcolor{blue}{88.5}} & 87.0 & 83.8 & 91.3 & \textcolor{blue}{64.6} & 39.4 & 90.7 & \textcolor{blue}{94.4}\\
SQuAD-2 & 47.1 & 91.9 & \textcolor{blue}{87.4} & \textcolor{blue}{87.2} & 87.1 & \textbf{\textcolor{blue}{84.6}} & \textbf{\textcolor{blue}{91.7}} & \textcolor{blue}{67.9} & 45.1 & \textbf{\textcolor{blue}{90.9}} & \textcolor{blue}{94.7}\\
NewsQA & 45.2 & 91.4 & \textcolor{blue}{86.9} & \textcolor{blue}{87.6} & 86.9 & 84.0 & 91.3 & \textcolor{blue}{63.2} & 36.6 & 90.4 & 93.9\\
HotpotQA & 43.3 & \textcolor{blue}{92.1} & \textbf{\textcolor{blue}{88.6}} & \textcolor{blue}{86.9} & 86.8 & 83.8 & 91.1 & \textcolor{blue}{66.1} & 39.4 & \textcolor{blue}{90.8} & \textcolor{blue}{94.2}\\
BoolQ & 51.0 & \textcolor{blue}{92.1} & \textcolor{blue}{86.3} & 85.8 & \textbf{\textcolor{blue}{87.4}} & 83.9 & 90.5 & 59.6 & 32.4 & 90.7 & 93.7\\
DROP & \textbf{\textcolor{blue}{53.4}} & \textcolor{blue}{92.3} & \textcolor{blue}{87.0} & \textcolor{blue}{87.9} & 87.1 & \textcolor{blue}{84.3} & 91.1 & \textbf{\textcolor{blue}{70.4}} & 42.3 & 90.7 & \textbf{\textcolor{blue}{94.9}}\\
\bottomrule
\end{tabular}
\end{adjustbox}
\caption{Out-of-class transfer results from question answering tasks to classification/regression tasks in the \smallsc{Full $\rightarrow$ Full} regime.}
\label{tbla7a}
\end{table}
\end{landscape}
\clearpage
\newpage
\global\pdfpageattr\expandafter{\the\pdfpageattr/Rotate 90}
\begin{landscape}
\begin{table}[t]
\centering
\begin{adjustbox}{max width=2\textwidth}
\begin{tabular}{ l l l l l  l l l l  l l l }
\toprule
\textbf{Task} & CoLA & SST-2 & MRPC & STS-B & QQP & MNLI & QNLI & RTE & WNLI & SNLI & SciTail \\
\midrule
\emph{Baseline} & \textcolor{orange}{4.7 $\pm$ 8.2} & \textcolor{orange}{77.5 $\pm$ 6.3} & \textcolor{orange}{81.9 $\pm$ 0.7} & \textcolor{orange}{29.9 $\pm$ 10.5} & \textcolor{orange}{25.7 $\pm$ 25.1} & \textcolor{orange}{38.7 $\pm$ 3.2} & \textcolor{orange}{62.4 $\pm$ 9.5} & \textcolor{orange}{54.7 $\pm$ 3.2} & \textcolor{orange}{44.4 $\pm$ 6.3} & \textcolor{orange}{46.7 $\pm$ 4.5} & \textcolor{orange}{64.1 $\pm$ 13.6}\\
SQuAD-1 & 3.3 $\pm$ 5.5 & \textcolor{blue}{83.3 $\pm$ 1.4} & \textcolor{blue}{83.4 $\pm$ 1.1} & \textbf{\textcolor{blue}{72.1 $\pm$ 7.0}} & \textcolor{blue}{47.8 $\pm$ 25.8} & \textcolor{blue}{49.4 $\pm$ 2.3} & \textbf{\textcolor{blue}{86.9 $\pm$ 0.6}} & \textcolor{blue}{63.1 $\pm$ 1.7} & 41.7 $\pm$ 7.4 & \textcolor{blue}{53.8 $\pm$ 2.1} & \textbf{\textcolor{blue}{86.6 $\pm$ 1.2}}\\
SQuAD-2 & 2.9 $\pm$ 6.5 & \textbf{\textcolor{blue}{83.8 $\pm$ 1.6}} & \textcolor{blue}{82.1 $\pm$ 0.8} & \textcolor{blue}{65.0 $\pm$ 12.1} & \textcolor{blue}{42.1 $\pm$ 32.1} & \textbf{\textcolor{blue}{52.0 $\pm$ 6.4}} & \textcolor{blue}{83.6 $\pm$ 2.2} & \textcolor{blue}{61.6 $\pm$ 2.2} & 44.4 $\pm$ 7.0 & \textbf{\textcolor{blue}{55.3 $\pm$ 4.1}} & \textcolor{blue}{67.9 $\pm$ 16.9}\\
NewsQA & 1.8 $\pm$ 3.8 & \textcolor{blue}{81.4 $\pm$ 3.1} & \textcolor{blue}{83.6 $\pm$ 1.3} & \textcolor{blue}{67.1 $\pm$ 6.4} & \textcolor{blue}{52.2 $\pm$ 25.1} & \textcolor{blue}{48.5 $\pm$ 3.2} & \textcolor{blue}{79.2 $\pm$ 6.0} & \textcolor{blue}{63.5 $\pm$ 3.0} & 43.5 $\pm$ 6.0 & \textcolor{blue}{54.3 $\pm$ 2.5} & \textcolor{blue}{83.0 $\pm$ 8.0}\\
\bottomrule
\end{tabular}
\end{adjustbox}
\caption{Out-of-class transfer results from question answering tasks to classification/regression tasks in the \smallsc{Full $\rightarrow$ Limited} regime.}
\label{tbla7b}
\end{table}
\end{landscape}
\clearpage
\newpage
\begin{landscape}
\begin{table}[t]
\centering
\begin{adjustbox}{max width=2\textwidth}
\begin{tabular}{ l l l l l  l l l l  l l l }
\toprule
\textbf{Task} & CoLA & SST-2 & MRPC & STS-B & QQP & MNLI & QNLI & RTE & WNLI & SNLI & SciTail \\
\midrule
\emph{Baseline} & \textcolor{orange}{4.7 $\pm$ 8.2} & \textcolor{orange}{77.5 $\pm$ 6.3} & \textcolor{orange}{81.9 $\pm$ 0.7} & \textcolor{orange}{29.9 $\pm$ 10.5} & \textcolor{orange}{25.7 $\pm$ 25.1} & \textcolor{orange}{38.7 $\pm$ 3.2} & \textcolor{orange}{62.4 $\pm$ 9.5} & \textcolor{orange}{54.7 $\pm$ 3.2} & \textcolor{orange}{44.4 $\pm$ 6.3} & \textcolor{orange}{46.7 $\pm$ 4.5} & \textcolor{orange}{64.1 $\pm$ 13.6}\\
SQuAD-1 & \textbf{\textcolor{blue}{8.6 $\pm$ 10.0}} & 74.7 $\pm$ 7.2 & 81.8 $\pm$ 0.7 & \textcolor{blue}{54.3 $\pm$ 8.9} & \textcolor{blue}{38.5 $\pm$ 28.6} & \textcolor{blue}{39.9 $\pm$ 3.2} & \textcolor{blue}{73.9 $\pm$ 2.1} & \textcolor{blue}{56.2 $\pm$ 3.5} & \textcolor{blue}{45.8 $\pm$ 7.6} & \textcolor{blue}{48.4 $\pm$ 3.6} & \textcolor{blue}{70.4 $\pm$ 13.7}\\
SQuAD-2 & \textcolor{blue}{7.5 $\pm$ 10.4} & 77.3 $\pm$ 5.8 & 81.8 $\pm$ 0.8 & \textcolor{blue}{51.4 $\pm$ 9.3} & \textcolor{blue}{38.4 $\pm$ 29.0} & \textcolor{blue}{41.8 $\pm$ 2.5} & \textcolor{blue}{74.8 $\pm$ 1.8} & \textcolor{blue}{56.9 $\pm$ 3.2} & \textcolor{blue}{45.0 $\pm$ 5.7} & \textcolor{blue}{49.3 $\pm$ 3.7} & \textcolor{blue}{71.4 $\pm$ 15.3}\\
NewsQA & 3.3 $\pm$ 5.6 & 76.6 $\pm$ 6.3 & \textcolor{blue}{82.0 $\pm$ 0.7} & \textcolor{blue}{59.4 $\pm$ 8.2} & \textcolor{blue}{45.7 $\pm$ 24.6} & \textcolor{blue}{42.3 $\pm$ 2.6} & \textbf{\textcolor{blue}{77.6 $\pm$ 1.1}} & \textcolor{blue}{59.0 $\pm$ 2.7} & 43.7 $\pm$ 7.5 & \textcolor{blue}{49.9 $\pm$ 2.6} & \textcolor{blue}{77.3 $\pm$ 11.2}\\
\bottomrule
\end{tabular}
\end{adjustbox}
\caption{Out-of-class transfer results from question answering tasks to classification/regression tasks in the \smallsc{Limited $\rightarrow$ Limited} regime.}
\label{tbla7c}
\end{table}
\end{landscape}
\clearpage
\newpage
\global\pdfpageattr\expandafter{\the\pdfpageattr/Rotate 90}
\begin{landscape}
\begin{table}[t]
\centering
\begin{adjustbox}{max width=2\textwidth}
\begin{tabular}{ l l l l l  l l l l  l l l }
\toprule
\textbf{Task} & CoLA & SST-2 & MRPC & STS-B & QQP & MNLI & QNLI & RTE & WNLI & SNLI & SciTail \\
\midrule
\emph{Baseline} & \textcolor{orange}{51.0} & \textcolor{orange}{91.9} & \textcolor{orange}{84.0} & \textcolor{orange}{85.9} & \textcolor{orange}{87.3} & \textcolor{orange}{84.2} & \textcolor{orange}{91.4} & \textcolor{orange}{60.6} & \textcolor{orange}{45.1} & \textcolor{orange}{90.7} & \textcolor{orange}{93.9}\\
CCG & 46.2 & 90.5 & 83.7 & \textcolor{blue}{86.3} & 86.4 & 83.4 & 90.2 & \textcolor{blue}{61.7} & 35.2 & 90.6 & 93.3\\
POS-PTB & 39.7 & 91.2 & \textcolor{blue}{85.7} & \textcolor{blue}{86.2} & 86.9 & 82.9 & 90.3 & \textcolor{blue}{61.7} & 42.3 & \textbf{\textcolor{blue}{90.8}} & 91.9\\
POS-EWT & 49.4 & \textcolor{blue}{92.0} & \textcolor{blue}{84.6} & \textbf{\textcolor{blue}{86.9}} & 87.2 & 84.1 & 90.9 & \textcolor{blue}{63.2} & \textbf{\textcolor{blue}{56.3}} & 90.6 & 92.9\\
Parent & 47.7 & 91.9 & \textcolor{blue}{84.7} & \textcolor{blue}{86.1} & 87.0 & 84.0 & 90.4 & \textbf{\textcolor{blue}{65.3}} & 35.2 & \textbf{\textcolor{blue}{90.8}} & 92.8\\
GParent & 49.9 & 91.7 & 83.5 & 85.9 & 86.9 & 84.0 & 89.9 & 60.3 & \textcolor{blue}{52.1} & 90.6 & 92.9\\
GGParent & 49.2 & 91.4 & \textcolor{blue}{84.3} & \textcolor{blue}{86.2} & 86.9 & 83.3 & 90.9 & 57.0 & 43.7 & 90.3 & 90.9\\
\bottomrule
\end{tabular}
\end{adjustbox}
\caption{Out-of-class transfer results from sequence labeling tasks to classification/regression tasks in the \smallsc{Full $\rightarrow$ Full} regime.}
\label{tbla8a}
\end{table}
\end{landscape}
\clearpage
\newpage
\global\pdfpageattr\expandafter{\the\pdfpageattr/Rotate 90}
\begin{landscape}
\begin{table}[t]
\centering
\begin{adjustbox}{max width=2\textwidth}
\begin{tabular}{ l l l l l  l l l l  l l l }
\toprule
\textbf{Task} & CoLA & SST-2 & MRPC & STS-B & QQP & MNLI & QNLI & RTE & WNLI & SNLI & SciTail \\
\midrule
\emph{Baseline} & \textcolor{orange}{4.7 $\pm$ 8.2} & \textcolor{orange}{77.5 $\pm$ 6.3} & \textcolor{orange}{81.9 $\pm$ 0.7} & \textcolor{orange}{29.9 $\pm$ 10.5} & \textcolor{orange}{25.7 $\pm$ 25.1} & \textcolor{orange}{38.7 $\pm$ 3.2} & \textcolor{orange}{62.4 $\pm$ 9.5} & \textcolor{orange}{54.7 $\pm$ 3.2} & \textcolor{orange}{44.4 $\pm$ 6.3} & \textcolor{orange}{46.7 $\pm$ 4.5} & \textcolor{orange}{64.1 $\pm$ 13.6}\\
CCG & 0.0 $\pm$ 0.0 & 57.6 $\pm$ 3.2 & 81.5 $\pm$ 0.6 & \textcolor{blue}{63.6 $\pm$ 10.4} & \textcolor{blue}{57.2 $\pm$ 14.9} & \textcolor{blue}{42.8 $\pm$ 2.0} & \textbf{\textcolor{blue}{73.2 $\pm$ 1.1}} & \textcolor{blue}{56.9 $\pm$ 2.6} & \textcolor{blue}{44.7 $\pm$ 7.7} & \textcolor{blue}{48.8 $\pm$ 3.7} & \textcolor{blue}{78.3 $\pm$ 10.3}\\
POS-PTB & 0.2 $\pm$ 1.7 & 56.9 $\pm$ 4.0 & \textbf{\textcolor{blue}{82.3 $\pm$ 0.6}} & \textbf{\textcolor{blue}{68.7 $\pm$ 12.0}} & \textcolor{blue}{52.9 $\pm$ 22.8} & \textcolor{blue}{42.9 $\pm$ 1.0} & \textcolor{blue}{73.0 $\pm$ 0.7} & \textcolor{blue}{58.1 $\pm$ 1.7} & 42.7 $\pm$ 7.3 & \textcolor{blue}{49.9 $\pm$ 2.3} & \textcolor{blue}{80.1 $\pm$ 8.2}\\
POS-EWT & 0.9 $\pm$ 1.9 & 62.6 $\pm$ 4.2 & 81.9 $\pm$ 0.4 & \textcolor{blue}{50.1 $\pm$ 12.3} & \textcolor{blue}{44.1 $\pm$ 27.1} & \textcolor{blue}{42.0 $\pm$ 2.5} & \textcolor{blue}{72.1 $\pm$ 1.0} & \textcolor{blue}{56.3 $\pm$ 3.1} & \textbf{\textcolor{blue}{46.2 $\pm$ 7.1}} & \textcolor{blue}{48.9 $\pm$ 3.7} & \textcolor{blue}{78.0 $\pm$ 11.1}\\
Parent & \textcolor{blue}{10.1 $\pm$ 5.6} & 58.7 $\pm$ 2.9 & \textcolor{blue}{82.0 $\pm$ 0.7} & \textcolor{blue}{51.9 $\pm$ 16.4} & \textcolor{blue}{51.3 $\pm$ 13.7} & \textcolor{blue}{43.5 $\pm$ 2.8} & \textcolor{blue}{72.2 $\pm$ 1.1} & \textbf{\textcolor{blue}{59.7 $\pm$ 2.4}} & 42.5 $\pm$ 6.6 & \textcolor{blue}{49.7 $\pm$ 2.9} & \textcolor{blue}{79.5 $\pm$ 9.0}\\
\bottomrule
\end{tabular}
\end{adjustbox}
\caption{Out-of-class transfer results from sequence labeling tasks to classification/regression tasks in the \smallsc{Full $\rightarrow$ Limited} regime.}
\label{tbla8b}
\end{table}
\end{landscape}
\clearpage
\newpage
\global\pdfpageattr\expandafter{\the\pdfpageattr/Rotate 90}
\begin{landscape}
\begin{table}[t]
\centering
\begin{adjustbox}{max width=2\textwidth}
\begin{tabular}{ l l l l l  l l l l  l l l }
\toprule
\textbf{Task} & CoLA & SST-2 & MRPC & STS-B & QQP & MNLI & QNLI & RTE & WNLI & SNLI & SciTail \\
\midrule
\emph{Baseline} & \textcolor{orange}{4.7 $\pm$ 8.2} & \textcolor{orange}{77.5 $\pm$ 6.3} & \textcolor{orange}{81.9 $\pm$ 0.7} & \textcolor{orange}{29.9 $\pm$ 10.5} & \textcolor{orange}{25.7 $\pm$ 25.1} & \textcolor{orange}{38.7 $\pm$ 3.2} & \textcolor{orange}{62.4 $\pm$ 9.5} & \textcolor{orange}{54.7 $\pm$ 3.2} & \textcolor{orange}{44.4 $\pm$ 6.3} & \textcolor{orange}{46.7 $\pm$ 4.5} & \textcolor{orange}{64.1 $\pm$ 13.6}\\
CCG & 0.2 $\pm$ 1.0 & 57.9 $\pm$ 4.5 & 81.4 $\pm$ 0.4 & 1.4 $\pm$ 16.1 & \textcolor{blue}{30.6 $\pm$ 28.1} & 36.4 $\pm$ 3.2 & \textcolor{blue}{65.0 $\pm$ 8.4} & 53.2 $\pm$ 4.0 & \textcolor{blue}{49.0 $\pm$ 6.1} & 39.0 $\pm$ 4.6 & 56.4 $\pm$ 9.6\\
POS-PTB & 0.3 $\pm$ 1.4 & 58.8 $\pm$ 3.7 & \textbf{\textcolor{blue}{82.1 $\pm$ 0.6}} & 0.9 $\pm$ 12.1 & \textcolor{blue}{29.3 $\pm$ 27.5} & 37.9 $\pm$ 3.6 & \textcolor{blue}{63.6 $\pm$ 8.1} & 54.6 $\pm$ 4.0 & \textcolor{blue}{45.7 $\pm$ 7.6} & 41.9 $\pm$ 5.0 & \textcolor{blue}{65.2 $\pm$ 13.8}\\
POS-EWT & 0.3 $\pm$ 0.7 & 60.0 $\pm$ 6.5 & 81.6 $\pm$ 0.5 & 4.1 $\pm$ 7.3 & 23.9 $\pm$ 25.3 & 38.1 $\pm$ 3.2 & \textcolor{blue}{68.0 $\pm$ 4.0} & \textcolor{blue}{56.6 $\pm$ 2.4} & \textcolor{blue}{46.1 $\pm$ 7.2} & 41.0 $\pm$ 4.3 & \textcolor{blue}{65.8 $\pm$ 14.1}\\
Parent & 0.0 $\pm$ 0.9 & 62.6 $\pm$ 5.8 & 81.5 $\pm$ 0.6 & 7.7 $\pm$ 20.9 & \textcolor{blue}{39.4 $\pm$ 29.1} & \textcolor{blue}{40.8 $\pm$ 3.2} & \textcolor{blue}{67.8 $\pm$ 5.9} & \textcolor{blue}{58.8 $\pm$ 3.1} & \textcolor{blue}{44.5 $\pm$ 7.7} & 45.2 $\pm$ 4.4 & \textbf{\textcolor{blue}{79.9 $\pm$ 7.7}}\\
\bottomrule
\end{tabular}
\end{adjustbox}
\caption{Out-of-class transfer results from sequence labeling tasks to classification/regression tasks in the \smallsc{Limited $\rightarrow$ Limited} regime.}
\label{tbla8c}
\end{table}
\end{landscape}
\clearpage
\newpage
\global\pdfpageattr\expandafter{\the\pdfpageattr/Rotate 90}
\begin{landscape}
\begin{table}[t]
\centering
\begin{adjustbox}{max width=2\textwidth}
\begin{tabular}{ l l l l l  l l l l  l l l }
\toprule
\textbf{Task} & SQuAD-1 & SQuAD-2 & NewsQA & HotpotQA & BoolQ & DROP & WikiHop & DuoRC-p & DuoRC-s & CQ & ComQA \\
\midrule
\emph{Baseline} & \textcolor{orange}{87.9} & \textcolor{orange}{71.9} & \textcolor{orange}{64.1} & \textcolor{orange}{67.9} & \textcolor{orange}{65.7} & \textcolor{orange}{22.4} & \textcolor{orange}{62.8} & \textcolor{orange}{50.6} & \textcolor{orange}{63.3} & \textcolor{orange}{30.5} & \textcolor{orange}{63.2}\\
CoLA & 87.8 & 70.1 & \textcolor{blue}{64.6} & \textcolor{blue}{68.2} & 64.9 & 22.3 & \textcolor{blue}{62.9} & \textcolor{blue}{51.0} & \textbf{\textcolor{blue}{63.8}} & 30.0 & 62.7\\
SST-2 & 87.7 & 71.3 & \textcolor{blue}{64.9} & \textcolor{blue}{68.3} & \textcolor{blue}{68.0} & 22.2 & \textcolor{blue}{63.1} & \textbf{\textcolor{blue}{51.1}} & 63.2 & 28.1 & 62.2\\
MRPC & 87.8 & 67.7 & 63.8 & 66.4 & \textcolor{blue}{66.4} & 22.4 & 62.5 & \textcolor{blue}{51.0} & 63.1 & 26.9 & 62.5\\
STS-B & 87.9 & 70.1 & 64.0 & 66.2 & 64.9 & 22.1 & \textbf{\textcolor{blue}{63.4}} & \textcolor{blue}{51.0} & 62.4 & 29.7 & 62.9\\
QQP & 87.9 & 71.5 & 64.0 & \textcolor{blue}{68.8} & 64.9 & 22.1 & \textcolor{blue}{63.2} & 50.5 & 62.0 & \textcolor{blue}{33.2} & 61.4\\
MNLI & 87.4 & \textcolor{blue}{72.8} & \textcolor{blue}{64.9} & \textcolor{blue}{68.7} & \textbf{\textcolor{blue}{69.8}} & \textcolor{blue}{22.7} & \textcolor{blue}{63.3} & \textcolor{blue}{50.7} & 62.6 & \textbf{\textcolor{blue}{35.5}} & 61.6\\
\bottomrule
\end{tabular}
\end{adjustbox}
\caption{Out-of-class transfer results from classification/regression tasks to question answering tasks in the \smallsc{Full $\rightarrow$ Full} regime.}
\label{tbla9a}
\end{table}
\end{landscape}
\clearpage
\newpage
\global\pdfpageattr\expandafter{\the\pdfpageattr/Rotate 90}
\begin{landscape}
\begin{table}[t]
\centering
\begin{adjustbox}{max width=2\textwidth}
\begin{tabular}{ l l l l l  l l l l  l l l }
\toprule
\textbf{Task} & SQuAD-1 & SQuAD-2 & NewsQA & HotpotQA & BoolQ & DROP & WikiHop & DuoRC-p & DuoRC-s & CQ & ComQA \\
\midrule
\emph{Baseline} & \textcolor{orange}{26.8 $\pm$ 6.0} & \textcolor{orange}{50.1 $\pm$ 0.1} & \textcolor{orange}{28.8 $\pm$ 4.9} & \textcolor{orange}{23.3 $\pm$ 4.0} & \textcolor{orange}{62.2 $\pm$ 0.0} & \textcolor{orange}{19.4 $\pm$ 11.8} & \textcolor{orange}{25.5 $\pm$ 4.7} & \textcolor{orange}{41.6 $\pm$ 1.1} & \textcolor{orange}{38.9 $\pm$ 3.3} & \textcolor{orange}{30.7 $\pm$ 2.5} & \textcolor{orange}{39.1 $\pm$ 1.2}\\
CoLA & 26.2 $\pm$ 6.3 & 50.0 $\pm$ 0.1 & \textcolor{blue}{30.4 $\pm$ 5.0} & \textcolor{blue}{24.2 $\pm$ 3.4} & 62.2 $\pm$ 0.0 & \textcolor{blue}{20.5 $\pm$ 13.8} & \textcolor{blue}{27.3 $\pm$ 3.7} & \textcolor{blue}{42.2 $\pm$ 1.2} & \textcolor{blue}{41.5 $\pm$ 3.3} & \textcolor{blue}{31.2 $\pm$ 1.5} & \textcolor{blue}{39.2 $\pm$ 1.2}\\
SST-2 & 22.4 $\pm$ 6.1 & 50.1 $\pm$ 0.0 & \textcolor{blue}{30.4 $\pm$ 3.9} & \textcolor{blue}{25.5 $\pm$ 4.6} & 62.2 $\pm$ 0.0 & \textcolor{blue}{34.7 $\pm$ 14.3} & \textcolor{blue}{26.7 $\pm$ 3.5} & \textcolor{blue}{41.8 $\pm$ 1.2} & \textcolor{blue}{39.8 $\pm$ 2.7} & \textcolor{blue}{30.9 $\pm$ 2.6} & 38.7 $\pm$ 1.5\\
MRPC & 23.4 $\pm$ 4.8 & 50.1 $\pm$ 0.0 & 25.7 $\pm$ 3.9 & 21.2 $\pm$ 2.1 & 62.2 $\pm$ 0.0 & 12.0 $\pm$ 11.1 & 23.9 $\pm$ 3.3 & 39.7 $\pm$ 1.7 & 35.0 $\pm$ 5.3 & \textcolor{blue}{31.8 $\pm$ 2.3} & 38.6 $\pm$ 1.1\\
STS-B & \textcolor{blue}{34.1 $\pm$ 4.2} & 50.0 $\pm$ 0.0 & 24.6 $\pm$ 2.1 & 23.3 $\pm$ 3.0 & 62.2 $\pm$ 0.0 & \textbf{\textcolor{blue}{40.1 $\pm$ 20.6}} & 23.7 $\pm$ 4.5 & 40.0 $\pm$ 1.8 & 35.5 $\pm$ 2.3 & \textcolor{blue}{30.8 $\pm$ 2.3} & 38.2 $\pm$ 1.0\\
\bottomrule
\end{tabular}
\end{adjustbox}
\caption{Out-of-class transfer results from classification/regression tasks to question answering tasks in the \smallsc{Full $\rightarrow$ Limited} regime.}
\label{tbla9b}
\end{table}
\end{landscape}
\clearpage
\newpage
\global\pdfpageattr\expandafter{\the\pdfpageattr/Rotate 90}
\begin{landscape}
\begin{table}[t]
\centering
\begin{adjustbox}{max width=2\textwidth}
\begin{tabular}{ l l l l l  l l l l  l l l }
\toprule
\textbf{Task} & SQuAD-1 & SQuAD-2 & NewsQA & HotpotQA & BoolQ & DROP & WikiHop & DuoRC-p & DuoRC-s & CQ & ComQA \\
\midrule
\emph{Baseline} & \textcolor{orange}{26.8 $\pm$ 6.0} & \textcolor{orange}{50.1 $\pm$ 0.1} & \textcolor{orange}{28.8 $\pm$ 4.9} & \textcolor{orange}{23.3 $\pm$ 4.0} & \textcolor{orange}{62.2 $\pm$ 0.0} & \textcolor{orange}{19.4 $\pm$ 11.8} & \textcolor{orange}{25.5 $\pm$ 4.7} & \textcolor{orange}{41.6 $\pm$ 1.1} & \textcolor{orange}{38.9 $\pm$ 3.3} & \textcolor{orange}{30.7 $\pm$ 2.5} & \textcolor{orange}{39.1 $\pm$ 1.2}\\
CoLA & \textcolor{blue}{27.3 $\pm$ 6.0} & 50.0 $\pm$ 0.2 & \textcolor{blue}{29.3 $\pm$ 4.6} & \textbf{\textcolor{blue}{23.8 $\pm$ 4.2}} & 62.2 $\pm$ 0.0 & 17.0 $\pm$ 10.4 & 25.0 $\pm$ 4.5 & \textbf{\textcolor{blue}{41.8 $\pm$ 1.1}} & \textbf{\textcolor{blue}{40.1 $\pm$ 3.1}} & \textcolor{blue}{31.2 $\pm$ 1.5} & \textcolor{blue}{39.3 $\pm$ 1.1}\\
SST-2 & 26.5 $\pm$ 6.0 & 50.1 $\pm$ 0.1 & \textcolor{blue}{29.0 $\pm$ 4.6} & \textcolor{blue}{23.4 $\pm$ 3.9} & 62.2 $\pm$ 0.0 & \textcolor{blue}{22.4 $\pm$ 12.7} & 25.5 $\pm$ 4.5 & 41.5 $\pm$ 1.0 & \textcolor{blue}{39.3 $\pm$ 3.2} & \textcolor{blue}{30.9 $\pm$ 1.9} & \textbf{\textcolor{blue}{39.4 $\pm$ 1.1}}\\
MRPC & 23.4 $\pm$ 4.5 & 50.0 $\pm$ 0.1 & 25.9 $\pm$ 3.6 & 21.2 $\pm$ 2.1 & 62.2 $\pm$ 0.0 & 18.8 $\pm$ 12.3 & 25.3 $\pm$ 4.5 & 41.2 $\pm$ 1.0 & 36.7 $\pm$ 3.9 & \textcolor{blue}{31.4 $\pm$ 2.3} & 38.7 $\pm$ 1.7\\
STS-B & 26.3 $\pm$ 4.5 & 50.1 $\pm$ 0.0 & 24.6 $\pm$ 2.5 & 21.5 $\pm$ 1.5 & 62.2 $\pm$ 0.0 & \textcolor{blue}{26.8 $\pm$ 15.4} & 24.0 $\pm$ 4.5 & 41.2 $\pm$ 1.2 & 36.7 $\pm$ 3.7 & \textcolor{blue}{31.5 $\pm$ 2.1} & 38.5 $\pm$ 1.3\\
QQP & 19.0 $\pm$ 4.0 & 50.0 $\pm$ 0.1 & 26.9 $\pm$ 2.8 & 22.4 $\pm$ 2.5 & 62.2 $\pm$ 0.0 & 19.2 $\pm$ 11.3 & 24.4 $\pm$ 4.8 & 41.4 $\pm$ 1.1 & 37.3 $\pm$ 2.9 & \textcolor{blue}{31.5 $\pm$ 2.1} & 38.7 $\pm$ 1.0\\
\bottomrule
\end{tabular}
\end{adjustbox}
\caption{Out-of-class transfer results from classification/regression tasks to question answering tasks in the \smallsc{Limited $\rightarrow$ Limited} regime.}
\label{tbla9c}
\end{table}
\end{landscape}
\clearpage
\newpage
\global\pdfpageattr\expandafter{\the\pdfpageattr/Rotate 90}
\begin{landscape}
\begin{table}[t]
\centering
\begin{adjustbox}{max width=2\textwidth}
\begin{tabular}{ l l l l l  l l l l  l l l }
\toprule
\textbf{Task} & SQuAD-1 & SQuAD-2 & NewsQA & HotpotQA & BoolQ & DROP & WikiHop & DuoRC-p & DuoRC-s & CQ & ComQA \\
\midrule
\emph{Baseline} & \textcolor{orange}{87.9} & \textcolor{orange}{71.9} & \textcolor{orange}{64.1} & \textcolor{orange}{67.9} & \textcolor{orange}{65.7} & \textcolor{orange}{22.4} & \textcolor{orange}{62.8} & \textcolor{orange}{50.6} & \textcolor{orange}{63.3} & \textcolor{orange}{30.5} & \textcolor{orange}{63.2}\\
CCG & 87.0 & 68.1 & 63.8 & 66.3 & 65.5 & 22.0 & 62.2 & 49.7 & 62.1 & 30.5 & 61.1\\
POS-PTB & 87.4 & 70.2 & 62.2 & 65.8 & 64.7 & 21.6 & 62.2 & 49.7 & \textcolor{blue}{63.5} & 28.4 & 62.8\\
POS-EWT & 85.9 & 66.7 & 62.6 & 66.2 & 65.4 & 22.0 & 62.6 & 50.2 & \textbf{\textcolor{blue}{63.8}} & \textcolor{blue}{33.7} & 61.5\\
Parent & 87.4 & 69.5 & \textcolor{blue}{64.4} & 67.9 & \textcolor{blue}{66.4} & 21.9 & \textcolor{blue}{63.1} & \textbf{\textcolor{blue}{51.3}} & 63.3 & \textcolor{blue}{34.3} & 62.3\\
GParent & 87.6 & 70.2 & 64.1 & 67.9 & \textcolor{blue}{65.8} & \textbf{\textcolor{blue}{22.7}} & 61.9 & 50.5 & 62.8 & \textbf{\textcolor{blue}{35.1}} & 62.2\\
GGParent & 87.7 & 71.0 & \textcolor{blue}{64.8} & 67.1 & \textbf{\textcolor{blue}{67.0}} & 21.8 & 62.1 & 50.6 & 61.8 & 28.8 & 63.1\\
\bottomrule
\end{tabular}
\end{adjustbox}
\caption{Out-of-class transfer results from sequence labeling tasks to question answering tasks in the \smallsc{Full $\rightarrow$ Full} regime.}
\label{tbla10a}
\end{table}
\end{landscape}
\clearpage
\newpage
\global\pdfpageattr\expandafter{\the\pdfpageattr/Rotate 90}
\begin{landscape}
\begin{table}[t]
\centering
\begin{adjustbox}{max width=2\textwidth}
\begin{tabular}{ l l l l l  l l l l  l l l }
\toprule
\textbf{Task} & SQuAD-1 & SQuAD-2 & NewsQA & HotpotQA & BoolQ & DROP & WikiHop & DuoRC-p & DuoRC-s & CQ & ComQA \\
\midrule
\emph{Baseline} & \textcolor{orange}{26.8 $\pm$ 6.0} & \textcolor{orange}{50.1 $\pm$ 0.1} & \textcolor{orange}{28.8 $\pm$ 4.9} & \textcolor{orange}{23.3 $\pm$ 4.0} & \textcolor{orange}{62.2 $\pm$ 0.0} & \textcolor{orange}{19.4 $\pm$ 11.8} & \textcolor{orange}{25.5 $\pm$ 4.7} & \textcolor{orange}{41.6 $\pm$ 1.1} & \textcolor{orange}{38.9 $\pm$ 3.3} & \textcolor{orange}{30.7 $\pm$ 2.5} & \textcolor{orange}{39.1 $\pm$ 1.2}\\
CCG & 15.0 $\pm$ 0.9 & 50.1 $\pm$ 0.0 & 19.9 $\pm$ 2.5 & 19.9 $\pm$ 1.8 & 62.2 $\pm$ 0.0 & \textcolor{blue}{59.7 $\pm$ 5.4} & 16.4 $\pm$ 1.7 & 36.0 $\pm$ 1.5 & 26.1 $\pm$ 7.7 & \textcolor{blue}{31.1 $\pm$ 2.9} & 37.8 $\pm$ 1.7\\
POS-PTB & 14.7 $\pm$ 0.8 & 50.1 $\pm$ 0.0 & 15.7 $\pm$ 2.5 & 15.4 $\pm$ 2.2 & 62.2 $\pm$ 0.0 & \textcolor{blue}{59.5 $\pm$ 3.4} & 17.8 $\pm$ 1.7 & 35.9 $\pm$ 2.0 & 16.6 $\pm$ 7.6 & 29.9 $\pm$ 2.0 & 37.2 $\pm$ 1.2\\
POS-EWT & 14.2 $\pm$ 1.1 & 50.1 $\pm$ 0.0 & 16.9 $\pm$ 2.4 & 17.2 $\pm$ 2.5 & 62.2 $\pm$ 0.0 & \textbf{\textcolor{blue}{60.0 $\pm$ 4.0}} & 22.9 $\pm$ 2.0 & 36.7 $\pm$ 2.5 & 20.4 $\pm$ 9.7 & \textcolor{blue}{32.2 $\pm$ 2.1} & 38.2 $\pm$ 1.4\\
Parent & 19.1 $\pm$ 3.8 & 50.1 $\pm$ 0.1 & 23.8 $\pm$ 1.7 & 22.5 $\pm$ 2.0 & 62.2 $\pm$ 0.0 & \textcolor{blue}{47.8 $\pm$ 14.9} & 22.2 $\pm$ 2.5 & 37.7 $\pm$ 1.7 & 29.8 $\pm$ 3.8 & \textcolor{blue}{32.0 $\pm$ 2.4} & 38.7 $\pm$ 1.4\\
GParent & 14.3 $\pm$ 0.7 & 50.1 $\pm$ 0.0 & 23.4 $\pm$ 2.2 & 19.6 $\pm$ 2.0 & 62.2 $\pm$ 0.0 & \textcolor{blue}{49.5 $\pm$ 19.0} & 19.0 $\pm$ 1.7 & 38.0 $\pm$ 2.0 & 26.1 $\pm$ 6.0 & \textcolor{blue}{31.6 $\pm$ 2.1} & 38.0 $\pm$ 1.1\\
GGParent & 13.7 $\pm$ 0.4 & 50.1 $\pm$ 0.0 & 23.5 $\pm$ 2.7 & 17.9 $\pm$ 2.5 & 62.2 $\pm$ 0.0 & \textcolor{blue}{38.1 $\pm$ 17.1} & 17.1 $\pm$ 1.4 & 37.9 $\pm$ 2.2 & 27.0 $\pm$ 6.1 & \textcolor{blue}{32.0 $\pm$ 2.2} & 37.8 $\pm$ 1.7\\
\bottomrule
\end{tabular}
\end{adjustbox}
\caption{Out-of-class transfer results from sequence labeling tasks to question answering tasks in the \smallsc{Full $\rightarrow$ Limited} regime.}
\label{tbla10b}
\end{table}
\end{landscape}
\clearpage
\newpage
\global\pdfpageattr\expandafter{\the\pdfpageattr/Rotate 90}
\begin{landscape}
\begin{table}[t]
\centering
\begin{adjustbox}{max width=2\textwidth}
\begin{tabular}{ l l l l l  l l l l  l l l }
\toprule
\textbf{Task} & SQuAD-1 & SQuAD-2 & NewsQA & HotpotQA & BoolQ & DROP & WikiHop & DuoRC-p & DuoRC-s & CQ & ComQA \\
\midrule
\emph{Baseline} & \textcolor{orange}{26.8 $\pm$ 6.0} & \textcolor{orange}{50.1 $\pm$ 0.1} & \textcolor{orange}{28.8 $\pm$ 4.9} & \textcolor{orange}{23.3 $\pm$ 4.0} & \textcolor{orange}{62.2 $\pm$ 0.0} & \textcolor{orange}{19.4 $\pm$ 11.8} & \textcolor{orange}{25.5 $\pm$ 4.7} & \textcolor{orange}{41.6 $\pm$ 1.1} & \textcolor{orange}{38.9 $\pm$ 3.3} & \textcolor{orange}{30.7 $\pm$ 2.5} & \textcolor{orange}{39.1 $\pm$ 1.2}\\
CCG & 12.3 $\pm$ 1.0 & 50.1 $\pm$ 0.0 & 15.0 $\pm$ 2.8 & 12.5 $\pm$ 2.6 & 62.2 $\pm$ 0.0 & \textbf{\textcolor{blue}{61.0 $\pm$ 2.0}} & 24.7 $\pm$ 1.4 & 40.5 $\pm$ 1.1 & 8.0 $\pm$ 7.1 & 30.7 $\pm$ 1.8 & 38.4 $\pm$ 0.9\\
POS-PTB & 12.4 $\pm$ 1.1 & 50.1 $\pm$ 0.0 & 19.0 $\pm$ 2.6 & 16.2 $\pm$ 2.6 & 62.2 $\pm$ 0.0 & \textcolor{blue}{59.9 $\pm$ 5.9} & 22.1 $\pm$ 2.2 & 39.0 $\pm$ 1.3 & 23.4 $\pm$ 9.6 & 30.0 $\pm$ 2.1 & 38.3 $\pm$ 1.4\\
POS-EWT & 12.1 $\pm$ 0.7 & 50.1 $\pm$ 0.0 & 20.3 $\pm$ 4.4 & 18.2 $\pm$ 2.9 & 62.2 $\pm$ 0.0 & \textcolor{blue}{53.8 $\pm$ 8.6} & 23.4 $\pm$ 2.2 & 39.7 $\pm$ 1.2 & 21.7 $\pm$ 10.7 & \textcolor{blue}{31.4 $\pm$ 2.0} & 38.2 $\pm$ 1.4\\
Parent & 14.5 $\pm$ 2.2 & 50.1 $\pm$ 0.0 & 21.4 $\pm$ 2.7 & 17.6 $\pm$ 2.2 & 62.2 $\pm$ 0.0 & \textcolor{blue}{56.9 $\pm$ 10.7} & 21.3 $\pm$ 2.5 & 38.8 $\pm$ 0.9 & 25.0 $\pm$ 9.4 & \textcolor{blue}{32.2 $\pm$ 2.2} & 38.3 $\pm$ 1.5\\
GParent & 21.1 $\pm$ 5.2 & 50.1 $\pm$ 0.0 & 23.4 $\pm$ 1.9 & 19.6 $\pm$ 2.1 & 62.2 $\pm$ 0.0 & \textcolor{blue}{54.6 $\pm$ 8.9} & 21.6 $\pm$ 2.6 & 38.9 $\pm$ 1.6 & 32.0 $\pm$ 3.2 & \textbf{\textcolor{blue}{32.9 $\pm$ 1.6}} & 38.9 $\pm$ 1.4\\
GGParent & \textbf{\textcolor{blue}{31.3 $\pm$ 8.0}} & 49.6 $\pm$ 0.9 & 25.5 $\pm$ 3.4 & 23.2 $\pm$ 4.5 & 62.2 $\pm$ 0.0 & \textcolor{blue}{36.9 $\pm$ 19.9} & 25.3 $\pm$ 2.3 & 40.2 $\pm$ 1.3 & 35.7 $\pm$ 2.2 & \textcolor{blue}{31.8 $\pm$ 2.6} & 39.0 $\pm$ 1.7\\
\bottomrule
\end{tabular}
\end{adjustbox}
\caption{Out-of-class transfer results from sequence labeling tasks to question answering tasks in the \smallsc{Limited $\rightarrow$ Limited} regime.}
\label{tbla10c}
\end{table}
\end{landscape}
\clearpage
\newpage
\global\pdfpageattr\expandafter{\the\pdfpageattr/Rotate 90}
\begin{landscape}
\begin{table}[t]
\centering
\begin{adjustbox}{max width=2\textwidth}
\begin{tabular}{ l l l l l  l l l l  l l l }
\toprule
\textbf{Task} & CCG & POS-PTB & POS-EWT & Parent & GParent & GGParent & ST & Chunk & NER & GED & Conj \\
\midrule
\emph{Baseline} & \textcolor{orange}{95.6} & \textcolor{orange}{96.7} & \textcolor{orange}{96.6} & \textcolor{orange}{95.4} & \textcolor{orange}{91.9} & \textcolor{orange}{89.5} & \textcolor{orange}{95.8} & \textcolor{orange}{97.1} & \textcolor{orange}{94.7} & \textcolor{orange}{46.6} & \textcolor{orange}{89.4}\\
CoLA & 95.5 & 96.7 & \textbf{\textcolor{blue}{96.7}} & 95.2 & 91.8 & 89.4 & 95.8 & 97.0 & 94.6 & 46.6 & \textcolor{blue}{89.8}\\
SST-2 & 95.6 & 96.7 & 96.6 & 95.3 & 91.8 & 89.4 & 95.8 & 97.0 & 94.6 & \textcolor{blue}{47.0} & \textcolor{blue}{89.9}\\
MRPC & 95.6 & 96.6 & 96.6 & 95.2 & 91.9 & 89.4 & 95.8 & 97.0 & 94.5 & \textcolor{blue}{47.0} & \textcolor{blue}{90.3}\\
STS-B & 95.4 & 96.7 & 96.6 & 95.2 & 91.4 & 89.2 & 95.8 & 97.0 & 94.3 & 46.5 & \textcolor{blue}{89.8}\\
QQP & 95.5 & 96.7 & \textbf{\textcolor{blue}{96.7}} & 95.1 & 91.7 & 89.3 & 95.8 & 97.1 & 94.6 & 46.4 & \textbf{\textcolor{blue}{90.4}}\\
MNLI & 95.4 & 96.7 & 96.6 & 95.1 & 91.9 & 89.0 & 95.7 & 97.1 & 94.6 & 46.6 & \textbf{\textcolor{blue}{90.4}}\\
QNLI & 95.5 & 96.7 & \textbf{\textcolor{blue}{96.7}} & 95.3 & 91.8 & \textbf{\textcolor{blue}{89.6}} & 95.8 & 97.0 & 94.7 & \textcolor{blue}{46.9} & \textcolor{blue}{89.5}\\
RTE & 95.5 & 96.7 & 96.6 & 95.3 & \textbf{\textcolor{blue}{92.0}} & 89.5 & 95.8 & 96.9 & 94.7 & \textbf{\textcolor{blue}{47.4}} & \textcolor{blue}{89.7}\\
WNLI & 95.5 & 96.7 & 96.5 & 95.4 & 91.8 & 89.5 & 95.8 & 97.0 & 94.5 & 46.3 & 89.4\\
SNLI & 95.5 & 96.7 & \textbf{\textcolor{blue}{96.7}} & 95.2 & 91.8 & 89.3 & 95.8 & 97.0 & 94.3 & 46.3 & \textcolor{blue}{89.7}\\
SciTail & 95.5 & 96.7 & \textbf{\textcolor{blue}{96.7}} & 95.2 & \textbf{\textcolor{blue}{92.0}} & 89.4 & 95.8 & 97.0 & 94.5 & 46.2 & \textcolor{blue}{89.5}\\
\bottomrule
\end{tabular}
\end{adjustbox}
\caption{Out-of-class transfer results from classification/regression tasks to sequence labeling tasks in the \smallsc{Full $\rightarrow$ Full} regime.}
\label{tbla11a}
\end{table}
\end{landscape}
\clearpage
\newpage
\global\pdfpageattr\expandafter{\the\pdfpageattr/Rotate 90}
\begin{landscape}
\begin{table}[t]
\centering
\begin{adjustbox}{max width=2\textwidth}
\begin{tabular}{ l l l l l  l l l l  l l l }
\toprule
\textbf{Task} & CCG & POS-PTB & POS-EWT & Parent & GParent & GGParent & ST & Chunk & NER & GED & Conj \\
\midrule
\emph{Baseline} & \textcolor{orange}{53.2 $\pm$ 1.6} & \textcolor{orange}{85.1 $\pm$ 0.9} & \textcolor{orange}{89.3 $\pm$ 0.4} & \textcolor{orange}{81.9 $\pm$ 0.9} & \textcolor{orange}{62.8 $\pm$ 1.3} & \textcolor{orange}{43.3 $\pm$ 1.7} & \textcolor{orange}{76.7 $\pm$ 0.9} & \textcolor{orange}{87.7 $\pm$ 0.5} & \textcolor{orange}{77.4 $\pm$ 1.5} & \textcolor{orange}{29.1 $\pm$ 1.3} & \textcolor{orange}{73.3 $\pm$ 1.6}\\
CoLA & 48.9 $\pm$ 3.2 & 83.9 $\pm$ 0.9 & 88.8 $\pm$ 0.6 & 79.9 $\pm$ 0.9 & \textcolor{blue}{63.2 $\pm$ 1.0} & 42.4 $\pm$ 1.5 & 75.2 $\pm$ 1.5 & 87.1 $\pm$ 0.6 & 77.3 $\pm$ 2.3 & 26.8 $\pm$ 2.1 & 71.6 $\pm$ 2.6\\
SST-2 & 50.0 $\pm$ 2.2 & 83.5 $\pm$ 1.1 & 88.4 $\pm$ 0.7 & 79.0 $\pm$ 1.0 & 61.8 $\pm$ 1.6 & 42.5 $\pm$ 2.4 & 75.7 $\pm$ 1.1 & 86.9 $\pm$ 0.7 & \textcolor{blue}{78.6 $\pm$ 1.8} & 27.3 $\pm$ 0.7 & 73.3 $\pm$ 1.4\\
MRPC & 53.0 $\pm$ 1.7 & 84.8 $\pm$ 0.9 & 89.3 $\pm$ 0.5 & 80.8 $\pm$ 1.3 & 62.3 $\pm$ 1.5 & 42.7 $\pm$ 1.7 & \textcolor{blue}{76.8 $\pm$ 0.8} & 87.4 $\pm$ 0.5 & 77.2 $\pm$ 2.5 & 27.9 $\pm$ 2.6 & 72.7 $\pm$ 1.8\\
STS-B & \textbf{\textcolor{blue}{56.6 $\pm$ 1.5}} & \textbf{\textcolor{blue}{86.6 $\pm$ 0.9}} & \textbf{\textcolor{blue}{90.4 $\pm$ 0.4}} & 81.9 $\pm$ 1.2 & 61.4 $\pm$ 1.7 & 42.0 $\pm$ 2.8 & \textbf{\textcolor{blue}{77.7 $\pm$ 0.9}} & \textbf{\textcolor{blue}{87.9 $\pm$ 0.6}} & 72.1 $\pm$ 4.4 & 29.1 $\pm$ 3.2 & 72.1 $\pm$ 2.8\\
QQP & 50.3 $\pm$ 3.3 & 83.6 $\pm$ 0.9 & 88.8 $\pm$ 0.5 & 80.3 $\pm$ 1.2 & 62.1 $\pm$ 1.2 & 43.2 $\pm$ 1.4 & 75.1 $\pm$ 1.2 & 86.7 $\pm$ 0.7 & \textcolor{blue}{78.8 $\pm$ 1.4} & 26.5 $\pm$ 1.1 & 71.5 $\pm$ 1.6\\
MNLI & 50.1 $\pm$ 1.8 & 82.5 $\pm$ 1.0 & 88.6 $\pm$ 0.5 & 79.6 $\pm$ 0.8 & 61.4 $\pm$ 1.2 & 41.9 $\pm$ 2.0 & 74.7 $\pm$ 1.4 & 86.5 $\pm$ 0.6 & \textcolor{blue}{79.8 $\pm$ 1.6} & 27.0 $\pm$ 0.6 & \textcolor{blue}{74.2 $\pm$ 1.4}\\
\bottomrule
\end{tabular}
\end{adjustbox}
\caption{Out-of-class transfer results from classification/regression tasks to sequence labeling tasks in the \smallsc{Full $\rightarrow$ Limited} regime.}
\label{tbla11b}
\end{table}
\end{landscape}
\clearpage
\newpage
\global\pdfpageattr\expandafter{\the\pdfpageattr/Rotate 90}
\begin{landscape}
\begin{table}[t]
\centering
\begin{adjustbox}{max width=2\textwidth}
\begin{tabular}{ l l l l l  l l l l  l l l }
\toprule
\textbf{Task} & CCG & POS-PTB & POS-EWT & Parent & GParent & GGParent & ST & Chunk & NER & GED & Conj \\
\midrule
\emph{Baseline} & \textcolor{orange}{53.2 $\pm$ 1.6} & \textcolor{orange}{85.1 $\pm$ 0.9} & \textcolor{orange}{89.3 $\pm$ 0.4} & \textcolor{orange}{81.9 $\pm$ 0.9} & \textcolor{orange}{62.8 $\pm$ 1.3} & \textcolor{orange}{43.3 $\pm$ 1.7} & \textcolor{orange}{76.7 $\pm$ 0.9} & \textcolor{orange}{87.7 $\pm$ 0.5} & \textcolor{orange}{77.4 $\pm$ 1.5} & \textcolor{orange}{29.1 $\pm$ 1.3} & \textcolor{orange}{73.3 $\pm$ 1.6}\\
CoLA & \textcolor{blue}{53.9 $\pm$ 1.4} & \textcolor{blue}{85.3 $\pm$ 0.8} & \textcolor{blue}{89.4 $\pm$ 0.5} & \textbf{\textcolor{blue}{82.3 $\pm$ 1.0}} & \textcolor{blue}{63.3 $\pm$ 1.5} & \textcolor{blue}{43.4 $\pm$ 1.7} & \textbf{\textcolor{blue}{77.8 $\pm$ 2.5}} & \textbf{\textcolor{blue}{87.8 $\pm$ 0.4}} & \textcolor{blue}{77.7 $\pm$ 2.6} & \textbf{\textcolor{blue}{29.3 $\pm$ 1.4}} & \textcolor{blue}{74.0 $\pm$ 1.4}\\
SST-2 & \textbf{\textcolor{blue}{54.0 $\pm$ 2.2}} & \textcolor{blue}{85.2 $\pm$ 0.9} & \textcolor{blue}{89.4 $\pm$ 0.4} & \textcolor{blue}{82.1 $\pm$ 1.0} & \textcolor{blue}{63.4 $\pm$ 2.0} & \textcolor{blue}{43.8 $\pm$ 1.9} & \textcolor{blue}{76.9 $\pm$ 0.8} & \textbf{\textcolor{blue}{87.8 $\pm$ 0.7}} & \textcolor{blue}{77.9 $\pm$ 1.9} & 28.9 $\pm$ 1.3 & \textcolor{blue}{74.2 $\pm$ 1.1}\\
MRPC & 51.9 $\pm$ 2.2 & 84.8 $\pm$ 1.6 & 89.0 $\pm$ 0.5 & 81.0 $\pm$ 1.2 & \textcolor{blue}{63.1 $\pm$ 1.4} & 43.3 $\pm$ 1.8 & 76.3 $\pm$ 1.0 & 87.6 $\pm$ 0.4 & 77.4 $\pm$ 2.2 & 28.6 $\pm$ 1.6 & \textcolor{blue}{73.7 $\pm$ 2.0}\\
STS-B & \textcolor{blue}{53.5 $\pm$ 2.7} & 85.1 $\pm$ 0.9 & \textcolor{blue}{89.5 $\pm$ 0.4} & 81.3 $\pm$ 1.6 & \textcolor{blue}{62.9 $\pm$ 2.0} & \textcolor{blue}{43.4 $\pm$ 2.1} & \textcolor{blue}{77.1 $\pm$ 0.8} & 87.6 $\pm$ 0.6 & \textcolor{blue}{77.8 $\pm$ 1.7} & 28.8 $\pm$ 2.2 & 72.7 $\pm$ 2.2\\
\bottomrule
\end{tabular}
\end{adjustbox}
\caption{Out-of-class transfer results from classification/regression tasks to sequence labeling tasks in the \smallsc{Limited $\rightarrow$ Limited} regime.}
\label{tbla11c}
\end{table}
\end{landscape}
\clearpage
\newpage
\global\pdfpageattr\expandafter{\the\pdfpageattr/Rotate 90}
\begin{landscape}
\begin{table}[t]
\centering
\begin{adjustbox}{max width=2\textwidth}
%
\end{adjustbox}
\caption{Out-of-class transfer results from question answering tasks to sequence labeling tasks in the \smallsc{Full $\rightarrow$ Full} regime.}
\label{tbla12a}
\end{table}
\end{landscape}
\clearpage
\newpage
\global\pdfpageattr\expandafter{\the\pdfpageattr/Rotate 90}
\begin{landscape}
\begin{table}[t]
\centering
\begin{adjustbox}{max width=2\textwidth}
%
\end{adjustbox}
\caption{Out-of-class transfer results from question answering tasks to sequence labeling tasks in the \smallsc{Full $\rightarrow$ Limited} regime.}
\label{tbla12b}
\end{table}
\end{landscape}
\clearpage
\newpage
\global\pdfpageattr\expandafter{\the\pdfpageattr/Rotate 90}
\begin{landscape}
\begin{table}[t]
\centering
\begin{adjustbox}{max width=2\textwidth}
%
\end{adjustbox}
\caption{Out-of-class transfer results from question answering tasks to sequence labeling tasks in the  \smallsc{Limited $\rightarrow$ Limited} regime.}
\label{tbla12c}
\end{table}
\end{landscape}
\clearpage
\newpage
\global\pdfpageattr\expandafter{\the\pdfpageattr/Rotate 0}

\end{document}